\documentclass[letterpaper]{article} 
\usepackage{aaai25}  
\usepackage{times}  
\usepackage{helvet}  
\usepackage{courier}  
\usepackage[hyphens]{url}  
\usepackage{graphicx} 
\usepackage{natbib}  
\usepackage{caption} 
\usepackage{algorithm}
\usepackage{algorithmic}

\usepackage{newfloat}
\usepackage{listings}
\DeclareCaptionStyle{ruled}{labelfont=normalfont,labelsep=colon,strut=off} 
\floatstyle{ruled}
\newfloat{listing}{tb}{lst}{}
\floatname{listing}{Listing}
\usepackage{times}
\usepackage{latexsym}
\usepackage{multirow}
\usepackage{booktabs}
\usepackage{graphicx}
\usepackage{amssymb}
\usepackage{makecell}
\usepackage{amsmath, bm}
\usepackage{longtable}
\usepackage{subfigure}
\usepackage{caption}
\usepackage{enumerate}
\usepackage{enumitem}

\usepackage{longtable}
\usepackage{xspace}
\usepackage{comment}

\usepackage{cuted}
\usepackage{url}

\usepackage{amssymb}

\usepackage{pifont}

\usepackage{amsmath}

\title{Enhancing Relation Extraction \\ via Supervised Rationale Verification and Feedback}

\author{
    Yongqi Li\textsuperscript{\rm 1}, Xin Miao\textsuperscript{\rm 1}, Shen Zhou\textsuperscript{\rm 1}, Mayi Xu\textsuperscript{\rm 1}, Yuyang Ren\textsuperscript{\rm 1,3}, Tieyun Qian\textsuperscript{\rm 1,2}\thanks{Corresponding author.}
}
\affiliations{
    \textsuperscript{\rm 1}School of Computer Science, Wuhan University, China\\
    \textsuperscript{\rm 2}Intellectual Computing Laboratory for Cultural Heritage, Wuhan University, China\\
    \textsuperscript{\rm 3}Research Institute of Nuclear Power Operation, China\\

    \{liyongqi, miaoxin, shenzhou, xumayi\}@whu.edu.cn, renyy@cnnp.com.cn, qty@whu.edu.cn
}

\usepackage{bibentry}

\begin{document}

\maketitle

\begin{abstract}
Despite the rapid progress that existing automated feedback methods have made in correcting the output of large language models (LLMs), these methods cannot be well applied to the relation extraction (RE) task due to their designated feedback objectives and correction manner.
To address this problem, we propose a novel automated feedback framework for RE, which presents \emph{a rationale supervisor to verify the rationale} and provides \emph{re-selected demonstrations as feedback} to correct the initial prediction.
Specifically, we first design a causal intervention and observation method \emph{to collect biased/unbiased rationales for contrastive training the rationale supervisor}.
Then, we present a verification-feedback-correction procedure to \emph{iteratively enhance LLMs' capability of handling the RE task}.
Extensive experiments prove that our proposed framework significantly outperforms existing methods.
\end{abstract}


\begin{links}
    \link{Code}{https://github.com/NLPGM/SRVF}
\end{links}

\section{Introduction}


The relation extraction (RE) task aims to extract the semantic relation between entities in the text, which is an important task in information extraction.
Unlike previous fine-tuning strategies based on small language models~\cite{wu-2019-rbert}, recent studies~\cite{wan-2023-gpt_re,ma-2023-llm_rank_for_IE} leverage the strong instruction understanding abilities and rich intrinsic knowledge of large language models (LLMs)~\cite{rlhf-2022-chatgpt,touvron-2023-llama2,anthropic-2022-claude} to enhance the performance of RE.

Despite their significant progress, LLM based methods may suffer from relation bias when performing relation extraction.
For example, given a sentence \textit{``data is derived from a study''}, where \textit{``data''} and \textit{``study''} form the \textit{``Entity-Origin''} relation, LLMs may be influenced by the pre-trained knowledge and have the stereotype that \textit{``data is the product that someone produces''}, thus making a biased relation prediction \textit{``Product-Producer''}, which ignores that the real producer is investigators (producer of the study).
Furthermore, existing LLM based RE methods focus on the pre-selection of in-context demonstrations~\cite{wan-2023-gpt_re,ma-2023-cot_explicit_llm_re} or instruction design~\cite{Zhang-2023-LLM_QA4RE} to improve the performance. The verification and feedback mechanism for correcting the biased prediction is still missing from current LLM based  RE research.

\begin{figure}[t!]
\centering
\includegraphics[width=0.4\textwidth]{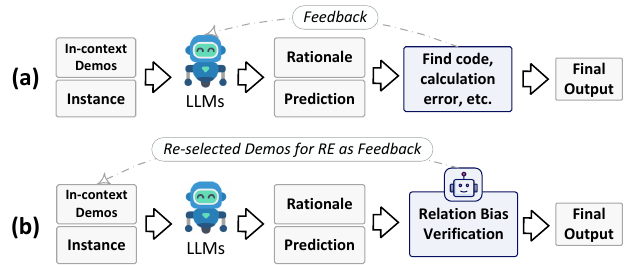}
\caption{Comparison between current automated feedback methods (a) and ours (b).
The main difference is that our rationale supervisor can verify whether the relation bias occurs and provide re-selected demonstrations as feedback.
}
\label{fig:introduction_figure}
\end{figure}


To fill this gap, in this study, we focus on exploring the verification and feedback mechanism~\cite{pan-2023-feedback_survey} of LLMs for RE.
Specifically, we aim to examine whether the relation prediction of LLMs is biased by verifying the rationale (the generated explanation when LLMs perform RE) and providing feedback for correction.
However, the current verification and feedback mechanism faces the following two problems when being applied to RE.

Firstly, existing methods are mainly designed for other tasks, e.g., the reasoning task. The objectives of their feedback are also tailored for those tasks, e.g., correcting code, factual, or calculation errors in initial responses~\cite{zhang-2023-algo,gou-2023-critic}, or choosing an optimal prefix for the next step in multi-step reasoning~\cite{khalifa-2023-grace},  as shown in Fig.~\ref{fig:introduction_figure} (a).
For example, for the mathematical reasoning task, Self-Refine~\cite{madaan-2023-self_refine} utilizes the LLM agent to find calculation errors in the initial answer and provide error information as feedback to correct the answer.
However, such feedback objectives are based on the logical properties of reasoning tasks, which are not available for RE.

Secondly, existing methods~\cite{madaan-2023-self_refine,nathani-2023-MAF} do not include demonstrations in their feedback. However, the demonstrations are essential for RE even at the correction stage. This is because without demonstrations in the feedback,  the RE task would degrade to zero-shot RE and is harder than the initial few-shot one. Moreover, the demonstrations in initial few-shot  RE cannot be directly used in feedback since they will mislead the model back to the initial one, and thus the impact of feedback is discarded.


To address the above problems, we propose a novel automated feedback framework for RE, which \emph{trains a rationale supervisor} based on a BERT-like small model and utilizes it to not only \emph{verify the prediction} but also \emph{provide new demonstration improved feedback for correction} during the inference. As shown in Fig.~\ref{fig:introduction_figure} (b), our rationale supervisor provides re-selected demonstrations as feedback for correcting the initial prediction of LLMs.

In order to train a rationale supervisor, we need to collect both unbiased and biased rationales, i.e., positive and negative samples. Though several verification methods have been proposed to collect positive and negative rationales in other tasks, both their purpose and the collection method are not suitable for our RE task.
(1) Firstly, their collected positive and negative rationales are used for training the verifier, which only needs to discriminate the positive predictions from negative ones.  In contrast, the rationale supervisor in our framework is designed to correct biased predictions, thus needing to further discriminate different negative rationales.
(2) Secondly, the way of collecting rationales in current verification methods relies on the manually annotated golden reasoning steps as positive samples and perform rule-based perturbation~\cite{paul-2023-REFINER,golovneva-2023-roscoe} or error step alignment~\cite{khalifa-2023-grace,li-2023-making_lm_with_step_aware_verifier} to obtain negative samples. Unfortunately, such annotated samples and rules for perturbation are not available in RE.

In view of this, we propose a causal intervention and observation method to address the lack of annotated rationales and collect biased rationales for training the supervisor. Specifically, we first present  \emph{a label-guided intervention strategy to collect unbiased rationales}, and we also present \emph{a diversified intervention strategy to collect biased rationales}.
In addition, during the inference, we utilize the rationale supervisor to retrieve new demonstrations from the labeled samples and include them in the feedback, which are then used by the LLM for re-generating predictions. Since the supervisor has learned the difference among various biased rationales, the LLM gets the signal to adjust its direction for correction. This verification-feedback-correction procedure iterates until the output rationale is verified as unbiased.


Overall, we make three major contributions. 1) We extend the LLM based RE research to the automated feedback paradigm, which equips LLM  with the ability of correcting the biased prediction. 2) We propose a novel supervised rationale verification and feedback framework, which first collects rationales with a causal intervention and observation method for training the supervisor, and then employs the supervisor to retrieve sample-related demonstrations as feedback for guiding the LLM in correction. 3) Extensive experiments prove that our proposed method can improve the performance of LLM based RE methods and is superior to existing automated feedback methods.

\section{Related Work}
\paragraph{LLMs for Relation Extraction}
Recently, many studies~\cite{xu-2023-LLM_IE_Survey,li-2023-evaluating_llms_for_IE_1,wei-2023-zero_shot_IE_chatgpt,wadhwa-2023-revisiting_RE_in_llms_era,li-2023-revisiting_llms_zs_RE} have explored how to unlock the potential of LLMs for the RE task, including designing the in-context demonstration selection strategy~\cite{wan-2023-gpt_re,ma-2023-cot_explicit_llm_re,pang-2023-guideline_for_icl_ie} and optimizing instruction patterns~\cite{Zhang-2023-LLM_QA4RE,wang-2023-llm_entity_bias,ma-2023-llm_rank_for_IE}.
Despite great success, these methods rely solely on optimizing the initial prompt to improve performance.
However, we find that due to the relation bias, LLMs may still confuse certain relations with similar entities and thus make biased predictions.
To alleviate this issue, we introduce the idea of automated feedback to RE for the first time, expecting to correct biased predictions via the provided feedback.

\paragraph{LLMs with Automated Feedback}
Some researchers have exploited the automated feedback for correcting the undesirable output of LLMs~\cite{pan-2023-feedback_survey,kamoi-2024-feedback_survey2}.
However, the feedbacks in existing methods are designed for correcting various reasoning mistakes, e.g., code errors~\cite{zhang-2023-algo}, factual errors~\cite{gou-2023-critic}, calculation errors~\cite{nathani-2023-MAF,madaan-2023-self_refine,paul-2023-REFINER}, or as an optimal prefix for the next step in multi-step reasoning~\cite{khalifa-2023-grace,li-2023-making_lm_with_step_aware_verifier}.
These feedbacks are dependent on the reasoning task and unavailable for RE. Moreover, they do not include the demonstrations which are essential for RE.
To address this issue, we propose a novel automated feedback framework which provides re-selected demonstrations as feedbacks to help LLMs correct the biased prediction.\looseness=-1

\section{Method}\label{sec:method}
This section presents our proposed supervised rationale verification and feedback (SRVF) framework for the RE task.
\paragraph{Task Formulation}
Given a set of pre-defined relation types $Y_D$, the relation extraction (RE) task aims to predict the relation type $y\in Y_D$ between the head entity $e^h$ and the tail entity $e^t$ of each test example $x=\{s, e^h, e^t\}$, where $s$ denotes the sentence.
In this study, we adopt in-context learning (ICL) with the rationale to prompt LLMs for the RE task.
Specifically, for each test example $x$, we need to randomly select or retrieve $m$ initial in-context demonstrations $D_{icl}=\{\{x_1,r_1^u,y_1\},...,\{x_m,r_m^u,y_m\}\}$ related to $x$ from the labeled dataset $D_{l}$~\footnote{Since there is no annotated golden rationale in the original dataset, we add the induced unbiased rationale in the following section to $D_l$ to enable it for the setup of ICL with the rationale, i.e., $D_l=\{\{x_1,r_1^u,y_1\},...,\{x_n,r_n^u,y_n\}\}$.}.
Then, the LLM $f_{\theta}$ with parameters $\theta$ is expected to output the relation type $y\in Y_D$ between $e^h$ and $e^t$, along with the rationale $r$, denoted as $\{r,y\} = f_{\theta}(D_{icl}, x)$.

\paragraph{Overview}
In this paper, we propose a rationale verification and feedback framework to guide LLMs towards better predictions for RE iteratively.
Generally, this framework consists of three phases: 1) causal intervention and observation for rationale collection, 2) contrastive training rationale supervisor, and 3) rationale verification and feedback.

Specifically, we first adopt the causal intervention and observation method to collect unbiased and biased rationales, i.e., $R_u$ and $R_{b}$.
Then, we use $R_u$ and $R_{b}$ to train the rationale supervisor $\mathcal{R_\gamma}$ with parameters $\gamma$.
Finally, as shown in Fig.~\ref{fig:framework}, in the inference time, once the output rationale $r$ is verified as a biased one by $\mathcal{R_\gamma}$, we use $\mathcal{R_\gamma}$ to retrieve feedback demonstrations $D_{fb}$ based on $r$, where $D_{fb} \subset D_{l}$.
The feedback demonstrations are used for re-generating $r$ and $y$ using ICL, i.e., $\{r,y\} = f_{\theta}(D_{fb}, x)$.
The procedure iterates until the rationale $r$ is verified as unbiased, and the corresponding relation prediction $y$ will finally be output.

\subsection{Causal Intervention and Observation for Rationale Collection}\label{sec:causal_intervention_observation}
\begin{figure}[t!]
\centering
\includegraphics[width=0.48\textwidth]{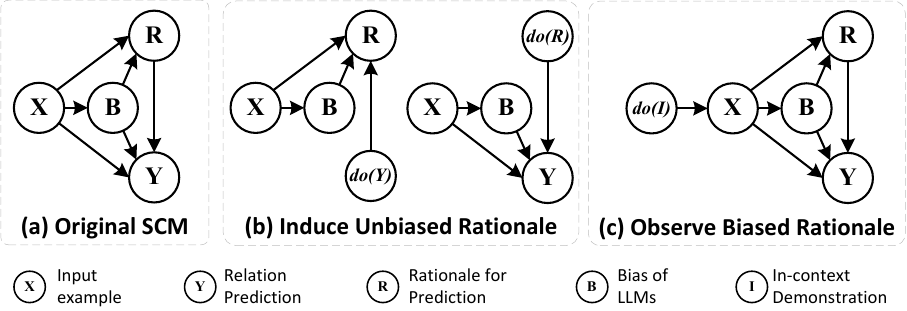}
\caption{
The structure causal model for illustrating the proposed causal intervention and observation strategy.
}
\label{fig:scm}
\end{figure}

Generally, during this phase, for each labeled sample $\{x_i,y_i\}$, we aim to collect the unbiased rationale corresponding with the golden label $\{r_i^u,y_i^u\}$, as well as the biased rationale with corresponding biased relation prediction $\{r_i^{b},y_i^{b}\}$.
This process consists of two steps: 1) induce unbiased rationale, and 2) observe biased rationale.
As shown in Fig.~\ref{fig:scm}, we use the structural causal model (SCM) in causal inference~\cite{pearl-2000-models_reasoning_inference} to illustrate the strategy.

\paragraph{Preliminary of SCM}
As shown in Fig.~\ref{fig:scm}, the SCMs show the relationships among the input ($X$), the relation prediction ($Y$), the rationale for prediction ($R$), the certain bias of LLMs ($B$) and in-context demonstration $I$.
The arrows between nodes indicate causal directions. For example, $``X\rightarrow R"$ means that the LLM generates the rationale $R$ related to the prediction for the sample $X$. $``X\rightarrow B\rightarrow R"$ indicates that the LLM activates some biased knowledge $B$ related to the sample $X$ and generates a rationale $R$ influenced by the biased knowledge $B$. 
Besides, in Fig.~\ref{fig:scm} (b), the ``$do(Y)$" indicates that cutting off all factors that could influence the value of $Y$ and assigning $Y$ a certain value as needed.

\paragraph{Induce Unbiased Rationale}
Previous methods rely on the human-annotated rationales, e.g., golden reasoning steps in mathematical tasks~\cite{khalifa-2023-grace}, which are not available in the RE dataset.
To address this issue, we propose \emph{a label-guided intervention strategy to obtain the unbiased rationale for each labeled sample}, which explains why the sample $x_i$ should be predicted as the golden label $y_i$.

As shown in Fig.~\ref{fig:scm} (b), this strategy consists of two steps:
1) cut causal directions that could make bias ($B$) influence the prediction ($Y$), and let the golden label guide the rationale ($R$) generation, formally denoted as $do(Y=y_i)$ and $do(Y)\rightarrow R$. The observed generated rationale is $R=r_i^u$;
2) conduct similar do-operation to the rationale $R$ and let $do(R)$ point to $Y$, i.e., $do(R=r_i^u), do(R)\rightarrow Y$. If the observed value of $Y$ is equal to the golden label $y_i$, we treat $\{r_i^u,y_i\}$ as the unbiased one and add it to $R_u$.

\paragraph{Observe Biased Rationale}
In previous methods, incorrect rationales are synthesized from golden ones using perturbation or error step alignment based on certain rules~\cite{golovneva-2023-roscoe,khalifa-2023-grace}.
However, these rules are designed based on the logical properties of reasoning tasks, which are not available in RE.
To tackle this problem, we propose \emph{a diversified intervention strategy for collecting the biased rationales}.

Specifically, for the labeled sample $\{x_i,y_i\}$, we first randomly select a demonstration set $D_{dii}$ with diverse labels, where $D_{dii}\subset D_{l}$ and the label of each demonstration in $D_{dii}$ is not equal to $y_i$.
The diversity of labels in $D_{dii}$ is designed to induce LLMs to make diverse errors on the same sample, to increase the diversity of collected biased rationales.
Then, as shown in Fig.~\ref{fig:scm} (c), we set the in-context demonstration $I$ as $\{x_j,r_j^u,y_j\}$ from $D_{dii}$, i.e., $do(I=\{x_j,r_j^u,y_j\})$.
Finally, the observed value of rationale $R$ is $r_{obs}$ while the observed value of rationale $Y$ is $y_{obs}$.
If $y_{obs}\neq y_i$, we treat the observed $r_{obs}$ with its corresponding relation prediction $y_{obs}$ as a potentially biased one, i.e., $\{r_i^{b},y_i^{b}\}$, and add it to $R_b$.

\subsection{Contrastive Training Rationale Supervisor}\label{sec:rationale_supervisor_training}
We expect the rationale supervisor to 1) verify whether the output rationale is biased, and 2) provide different feedbacks for different bias situations to correct the initial prediction.
To reach this, we adopt contrastive learning to train the rationale supervisor to acquire two abilities: 1) discriminating biased and unbiased rationales, and 2) learning the difference of various biased rationales.

\begin{figure*}[tbh!]
\centering
\includegraphics[width=0.8\textwidth]{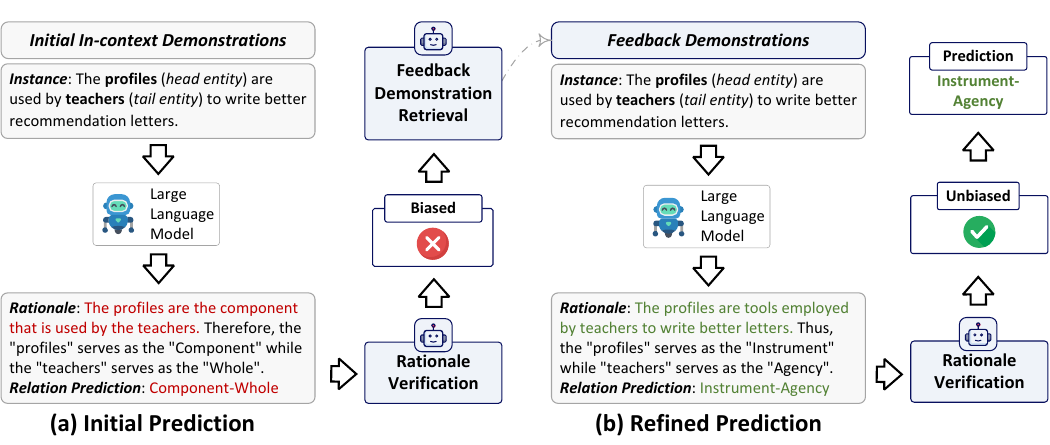}
\caption{
An example of correcting the initial biased prediction of LLMs via the proposed SRVF framework in the inference time.
The rationale supervisor first verifies the initial prediction in (a) as biased. Then, with the feedback demonstrations retrieved by the rationale supervisor, the LLM makes a correct relation prediction in (b).
Note: The rationale supervisor \includegraphics[scale=0.013]{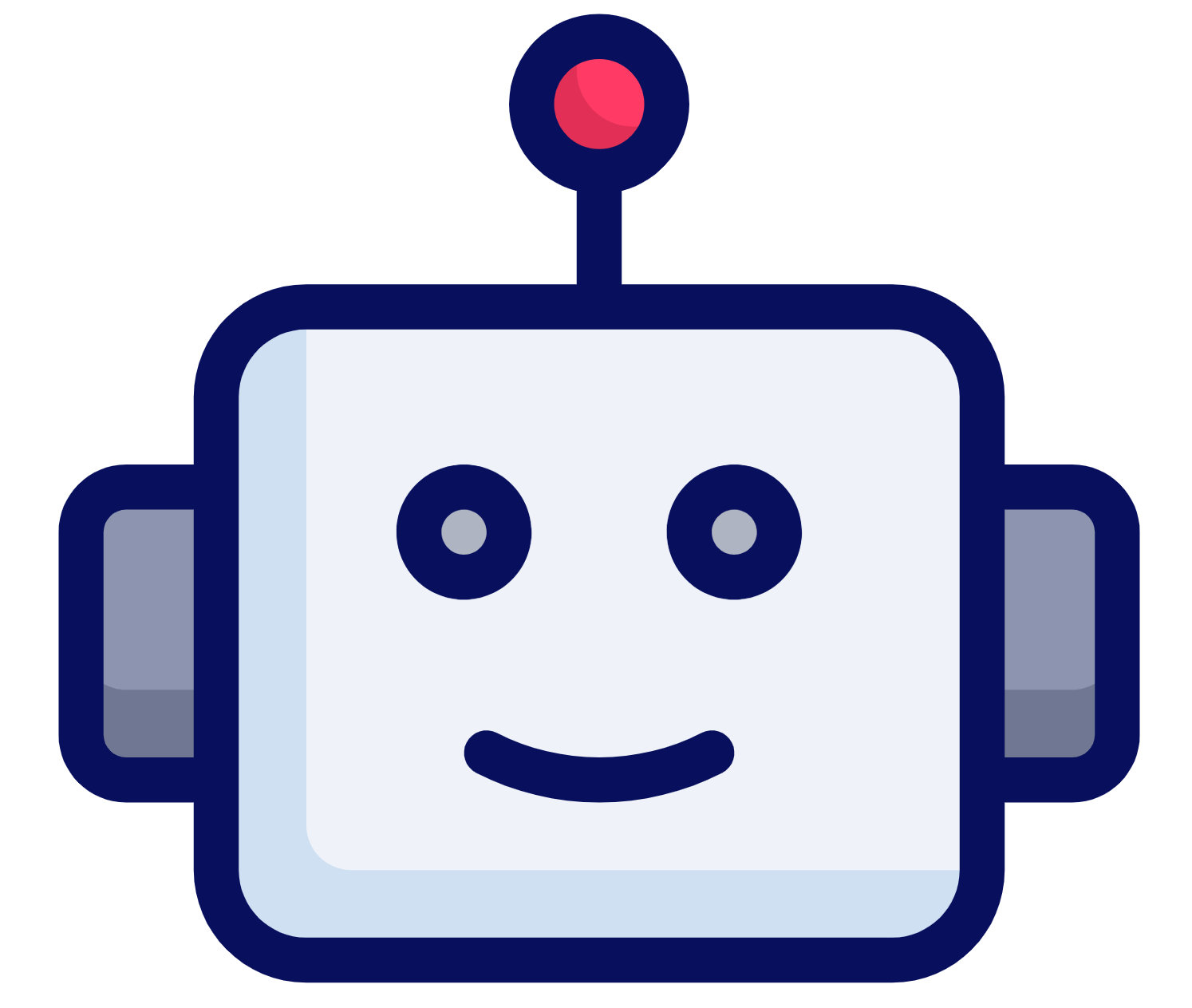} here is obtained by contrastive training using collected biased and unbiased rationales as described before.
}
\label{fig:framework}
\end{figure*}

We design two kinds of positive and negative pairs for contrastive training.

For positive pairs, we treat ``unbiased rationales with the same golden label'', and ``biased rationales under the same bias situation'' as the two kinds of positive pairs.
For example, if samples $s_1$ and $s_2$, which have the same label, are also predicted as the same wrong relation, we call ``samples $s_1$ and $s_2$ are in the same bias situation''.
Thus, the biased rationales ($r_1^b$ and $r_2^b$) of $s_1$ and $s_2$, are treated as a positive pair and should be pulled together in the rationale representation space, i.e., $r_1^b \rightarrow \leftarrow r_2^b$.

For negative pairs, we first consider the ``biased and unbiased rationales from the same sample'' as a negative pair.
This is designed to train the rationale supervisor to distinguish between biased and unbiased rationales.
For example, a sample $s_1=\{r_1^u,y_1\}$ where $y_1$ is the golden label and $r_1^u$ is the corresponding unbiased rationale, is wrongly predicted as relation $y_2$ and corresponding biased rationale is $r_1^b$.
Thus, $r_1^b$ and $r_1^u$ are treated as a negative pair and should be pushed away in the rationale representation space, i.e., $r_1^b\leftrightarrow r_1^u$.
Second, we also treat ``biased rationales under different bias situations'' as a negative pair to train the rationale supervisor, which can distinguish different bias situations and provide feedback based on the biased rationale in the inference time.

In general, the contrastive loss is calculated as:

\newcommand{\clposset}{{{S}^{pos}}\xspace}
\newcommand{\clnegset}{{{S}^{neg}}\xspace}

\begin{equation}\label{eq:rationale_contrastive_loss}
\mathcal{L}_{cl} = -\log
    \frac
    {
    \frac{1}{\|\clposset \|}\sum\limits_{\{r_1,r_2\}\in \clposset}
        \exp(sim(r_1,r_2)/\tau)
    }
    {
        \sum\limits_{\{r_1,r_2\}\in (\clposset \cup \clnegset)}
        \exp(sim(r_1,r_2)/\tau)
    },
\end{equation}

\begin{equation}\label{eq:sim_function}
sim(r_1,r_2)=\mathcal{R_\gamma}(r_1)\cdot \mathcal{R_\gamma}(r_2)^{\mathsf{T}},
\end{equation}

where $\clposset = {{S}^{pos}_1} \cup {{S}^{pos}_2},~ \clnegset = {{S}^{neg}_1} \cup {{S}^{neg}_2}$. ${{S}^{pos}_1}$ and ${{S}^{pos}_2}$ denote the two kinds of positive pair set, and ${{S}^{neg}_1}$ and ${{S}^{neg}_2}$ denote two kinds of negative pair set.
Here we adopt the dot product as the similarity function $sim()$ and add a temperature hyper-parameter $\tau$ to focus more on difficult pairs~\cite{chen-2020-simclr}.
During the procedure of rationale contrastive training, the parameters $\gamma$ of $\mathcal{R_\gamma}$ are updated to minimize $\mathcal{L}_{cl}$.

\subsection{Rationale Verification and Feedback}
As shown in Fig.~\ref{fig:framework}, in the inference time, the trained rationale supervisor $\mathcal{R_\gamma}$ first verifies whether the prediction is biased.
If the prediction is biased, the rationale supervisor will retrieve a feedback demonstration set, which then guides LLMs toward refined predictions.
In this subsection, we will elaborate on the ``\textit{Rationale Verification}'' and ``\textit{Feedback Demonstration Retrieval}'' in Fig.~\ref{fig:framework} in detail.
Here we denote the test example, output rationale, and relation prediction of LLMs as $x$, $r$, and $y$, respectively.

\paragraph{Rationale Verification}
For verification, we need to select the subsets $S_b$ and $S_u$ related to the prediction $y$ from $R_b$ and $R_u$, respectively, which are then used as anchors to determine whether the current output rationale is close to the biased or unbiased groups.
$S_b$ and $S_u$ are defined as follows:

\begin{equation}
S_b = \{\{r^b,y^b\}~|~\{r^b,y^b\} \in R_b, y^b=y\},
\end{equation}
\begin{equation}
S_u = \{\{r^u,y^u\}~|~\{r^u,y^u\} \in R_u, y^u=y\},
\end{equation}

Then, the indicator score to judge whether $r$ is a biased rationale is calculated as follows:

\begin{equation}\label{eq:rationale_verification_b}
p_b = {\max\limits_{\{r^b,y^b\} \in S_b} sim(r,r^b)} - {\max\limits_{\{r^u,y^u\} \in S_u} sim(r,r^u)},
\end{equation}

where the similarity function $sim()$ is defined in Eq.~\ref{eq:sim_function}.
When $p_b$ is greater than 0, it implies that the feature of $r$ is closer to the feature field of $S_b$ than that of $S_u$, which means $r$ and corresponding relation prediction $y$ should be regarded as biased, and feedback is needed to correct them.

\paragraph{Feedback Demonstration Retrieval}
Once the output rationale $r$ is verified as biased, we need to retrieve a new set of in-context demonstrations based on the feature of $r$ for guiding LLMs toward correct predictions.
Specifically, we first select the $k$ most similar biased rationales to $r$ in $S_b$, denoted as $S_b^{topk}$, which is defined as:
\begin{equation}
S_b^{topk} = \{\{r^b,y^b\}~|~rank_{\{r^b,y^b\} \in S_b}(sim(r, r^b)) \leq k \},
\end{equation}
Then, we select the labeled samples corresponding to the biased rationales in $S_b^{topk}$ from $D_l$ as the feedback demonstrations $D_{fb}$, which is defined as:
\begin{equation}
D_{fb} = \{\{x_i,r_i^u,y_i\}~|~\{x_i,r_i^u,y_i\} \in D_l, \{r_i^b,y_i^b\} \in S_b^{topk}\},
\end{equation}

where the biased $\{r_i^b,y_i^b\}$ and unbiased $\{r_i^u,y_i\}$ correspond to the same labeled sample $\{x_i,y_i\}$.

\paragraph{Correction via In-context Learning}
After the feedback demonstrations $D_{fb}$ are selected, we re-generate $r$ and $y$ using the LLM $f_\theta$, i.e., $\{r,y\} = f_{\theta}(D_{fb}, x)$.
This process will be iteratively performed until $r$ is verified as unbiased, and the corresponding prediction $y$ will be finally output.

\begin{table*}[t]
    \centering
    
    \setlength{\tabcolsep}{0.5mm}
    {
        \begin{tabular}{llccccccccccccc}
        \toprule
        {\multirow{2}{*}{\textbf{}}} & {\multirow{2}{*}{\textbf{Method}}} & \multicolumn{4}{c}{\textbf{SemEval}} & \multicolumn{4}{c}{\textbf{TACRED}} & \multicolumn{4}{c}{\textbf{Re-TACRED}} & {\multirow{2}{*}{\textbf{Avg.}}}\\
        \cmidrule(lr){3-6} \cmidrule(lr){7-10} \cmidrule(lr){11-14}
        \multicolumn{2}{c}{}                                 & 5-shot  & 10-shot  & 20-shot & 50-shot & 5-shot  & 10-shot & 20-shot & 50-shot & 5-shot  & 10-shot & 20-shot & 50-shot    \\
        \midrule
        
        \multirow{5}{*}{\rotatebox{90}{Random}}    & In-context Learning            & 48.40 & 49.11 & 49.65 & 49.31 & 24.17 & 23.69 & 24.66 & 24.21 & 21.37 & 21.99 & 21.52 & 21.10 & 31.60\\
                                   & ~w/ Self-Refine                                & 47.93 & 48.50 & 49.19 & 48.88 & 23.15 & 23.05 & 24.18 & 23.12 & 21.04 & 21.51 & 20.71 & 21.32 & 31.05\\                            
                                   & ~w/ Self-Consistency                           & 49.30 & 49.09 & 50.11 & 50.35 & 25.69 & 24.76 & 25.64 & 25.42 & 22.08 & 22.56 & 21.84 & 22.01 & 32.40\\                            
                                   & ~w/ GRACE                                      & 50.80 & 49.22 & 54.28 & 54.83 & 25.89 & 25.78 & 26.49 & 26.46 & 22.50 & 22.67 & 23.65 & 24.34 & 33.91\\
                                   \cmidrule(lr){2-15}
                                   
                                   & \textbf{~w/ our SRVF}   & \textbf{54.89} & \textbf{59.67} & \textbf{62.98} & \textbf{71.27} & \textbf{30.07} & \textbf{31.42} & \textbf{32.84} & \textbf{34.58} & \textbf{28.36} & \textbf{31.49} & \textbf{32.87} & \textbf{36.52} & \textbf{42.25}\\
                                   
        \midrule

        \multirow{5}{*}{\rotatebox{90}{SimCSE}}    & In-context Learning           & 57.33 & 59.13 & 62.49 & 64.26 & 27.48 & 28.64 & 30.08 & 27.81 & 34.78 & 41.85 & 42.82 & 43.69 & 43.36\\
                                   & ~w/ Self-Refine                               & 57.01 & 58.91 & 62.27 & 63.89 & 27.13 & 26.89 & 29.11 & 27.30 & 34.33 & 41.87 & 42.30 & 43.16 & 42.85\\
                                   & ~w/ Self-Consistency                          & 57.54 & 58.81 & 62.98 & 65.00 & 28.82 & 29.85 & 30.98 & 25.42 & 35.83 & 42.84 & 43.71 & 44.59 & 43.86\\
                                   & ~w/ GRACE                                     & 57.93 & 58.48 & 66.32 & 67.48 & 28.76 & 28.60 & 30.03 & 26.46 & 33.95 & 41.53 & 42.35 & 44.37 & 43.86\\
                                   \cmidrule(lr){2-15}                                   
                                   & \textbf{~w/ our SRVF}   & \textbf{60.76} & \textbf{64.12} & \textbf{69.54} & \textbf{76.32} & \textbf{32.99} & \textbf{33.50} & \textbf{34.81} & \textbf{36.13} & \textbf{39.48} & \textbf{46.54} & \textbf{49.73} & \textbf{54.31} & \textbf{49.85}\\
        
        \midrule
        
        \multirow{5}{*}{\rotatebox{90}{Task-specific}}    & In-context Learning    & 58.68 & 64.90 & 65.67 & 77.32 & 26.11 & 26.35 & 31.15 & 33.35 & 42.75 & 45.53 & 52.89 & 56.22 & 48.41\\
                                          & ~w/ Self-Refine                        & 58.38 & 64.96 & 65.68 & 77.35 & 25.01 & 25.48 & 30.62 & 32.67 & 42.10 & 44.98 & 52.11 & 55.62 & 47.91\\
                                          & ~w/ Self-Consistency                   & 59.62 & 65.45 & 65.74 & 77.48 & 26.93 & 26.83 & 31.67 & 33.61 & 43.54 & 46.04 & 53.34 & 56.69 & 48.91\\
                                          & ~w/ GRACE                              & 60.83 & 65.14 & 66.21 & 76.98 & 27.12 & 26.34 & 30.95 & 33.40 & 43.12 & 45.23 & 52.61 & 55.83 & 48.65\\
                                          \cmidrule(lr){2-15}                                          
                                          & \textbf{~w/ our SRVF}   & \textbf{62.12} & \textbf{67.03} & \textbf{68.94} & \textbf{80.08} & \textbf{30.50} & \textbf{30.92} & \textbf{34.83} & \textbf{36.32} & \textbf{46.13} & \textbf{48.09} & \textbf{55.07} & \textbf{59.82} & \textbf{51.65}\\

        \bottomrule
        \end{tabular}
    }
    \caption{Results (micro-F1 scores) on the SemEval, TACRED, and Re-TACRED datasets under various few-shot settings. Here we adopt the Llama-2-7b-chat as the LLM. The best results are in \textbf{bold}.}
    \label{tab:main_results_1}
\end{table*}

\section{Experiments}

\subsection{Evaluation Protocal}
\paragraph{Datasets and Metric}
We adopt three commonly used datasets for RE, including SemEval~\cite{hendrickx-2010-SemEval}, TACRED~\cite{zhang-2017-TACRED}, and Re-TACRED~\cite{stoica-2021-Re-TACRED}.
Besides, compared to the scenario with full data, the potential of LLMs under few-shot settings is of more concern~\cite {ma-2023-llm_rank_for_IE,xu-2023-LLM_IE_Survey}.
Hence we adopt the $k$-shot ( $k\in \{5,10,20,50\}$) settings to validate the effectiveness of the proposed method.
For all experiments, we report micro-F1 scores where \texttt{Other} and \texttt{no\_relation} are considered negative labels.

\paragraph{Backbones}
We experiment with three different methods as backbones for selecting initial in-context demonstrations for LLM based RE, including: 
1) \textit{Random}, which randomly selects initial demonstrations without any retriever.
2) \textit{SimCSE}, which uses SimCSE~\cite{gao-2021-simcse} to retrieve samples that have similar sentence semantics with the test example as initial in-context demonstrations.
3) \textit{Task-specific}, which uses a task-specific retriever that has been trained on the labeled samples~\cite{wan-2023-gpt_re}.

\paragraph{Baselines}
To the best of our knowledge, we are the first to explore the verification and feedback mechanism for LLM based RE.
Thus, we can only make modifications on current feedback methods in other tasks to adapt them for RE.
Specifically, we choose the following baselines:
\begin{itemize}
    \item \textit{Self-Refine}~\cite{madaan-2023-self_refine} consists of three LLM based agents, i.e., RE agent, verifier agent, and refiner agent, for iterative feedback and refinement.
    \item \textit{Self-Consistency}~\cite{wang-2023-selfconsistency} is proposed to conduct verification for the multiple candidate responses and choose the best response by majority voting.
    \item \textit{GRACE}~\cite{khalifa-2023-grace} trains a verifier to select the best intermediate reasoning step, which is then used as feedback for generating the next step.
\end{itemize}
For Self-Consistency, GRACE, and ours, the number of iterations or candidate responses is set to 5 for fairness.
For Self-Refine, the iteration number is set to 1 since we find that more iteration rounds result in performance degradation~\footnote{Please refer to Appendix for detailed implementation details of baselines and ours.}.

\subsection{Main Results}
Table~\ref{tab:main_results_1} reports the experimental results with various initial demonstration selection strategies on \textit{Llama-2-7b-chat} on the SemEval, TACRED, and Re-TACRED datasets.
From Table~\ref{tab:main_results_1}, we can draw the following conclusions:
1) Our proposed SRVF framework yields significant enhancements upon various backbones with different demonstration selection strategies.
Specifically, the improvement is most significant when randomly selecting the initial demonstrations, getting a 10.65\% absolute micro-F1 score increase on average.
Besides, when using SimCSE and task-specific retriever as backbones to carefully select initial in-context demonstrations, there are also 6.49\% and 3.24\% absolute micro-F1 score boosts on average, respectively.
2) Our proposed method exhibits significant superiority over existing verification and feedback methods under all settings.
The multi-agent based Self-Refine method is the worst, which is mainly due to its unsuitable feedback objectives and correction manner.
Existing methods for verifying the output of LLMs, i.e., Self-Consistency and GRACE, can enhance the performance of in-context learning to some extent. 
However, since they do not provide explicit feedback signals for LLMs to correct the prediction, their improvements are limited.

\subsection{Ablation Study}
To validate the effectiveness of components in our method, we introduce the following variants for ablation studies:
\begin{itemize}
    \item \textit{w/o label-guided intervention (LGI)}, where the labels do not guide the collecting of unbiased rationales.
    \item \textit{w/o diversified intervention (DI)}, which replaces the DI with random sampling for collecting biased rationales.
    \item \textit{w/o rational contrastive training (RCT)}, which trains the rationale supervisor with cross-entropy loss.
    \item \textit{w/o feedback demonstration retrieval (FDR)}, which removes the FDR strategy and uses the initially selected demonstrations as the feedback.
    \item \textit{w/o RG}, which skips the re-generation process and directly adopts the label of the top-1 retrieved demonstration as the final prediction.
\end{itemize}

The results of the ablation study are shown in Table~\ref{tab:ablation}. From the table, we make the following observations.
1) Removing LGI and DI strategies significantly degrades performance, indicating that LLMs struggle to collect unbiased rationales based solely on generation without causal intervention.
2) Eliminating RCT also reduces performance, demonstrating its effectiveness in helping the rationale supervisor distinguish between unbiased and various biased situations.
3) Omitting FDR significantly decreases performance, highlighting its crucial role in guiding LLMs toward corrected predictions despite iterative verification.
4) Removing the re-generation process results in a substantial performance drop, showcasing that simple assignment of retrieved top-1 demonstrations isn't sufficient and that in-context feedback for re-generation adds robustness to the correction process.

\begin{table}[t]
\small
    \centering
    \setlength{\tabcolsep}{0.6mm}
    {
    \begin{tabular}{lccccccc}
    \toprule
        \multirow{2}{*}{\textbf{Method}}  & \multicolumn{2}{c}{\textbf{SemEval}} & \multicolumn{2}{c}{\textbf{TACRED}} & \multicolumn{2}{c}{\textbf{Re-TACRED}} & \multirow{2}{*}{\textbf{Avg.}}\\
        \cmidrule(lr){2-3} \cmidrule(lr){4-5}\cmidrule(lr){6-7}
        & {{5-shot}} & {{10-shot}}  & {{5-shot}} & {{10-shot}}& {{5-shot}} & {{10-shot}}\\
        \midrule
        \textbf{Our SRVF} & \textbf{59.26} & \textbf{63.61} & \textbf{31.19} & \textbf{31.95} & \textbf{37.99} & {42.04} & \textbf{44.34} \\
        \midrule
        \textit{~w/o LGI} & 55.31 & 55.46 & 25.13 & 29.83 & 24.44 & 30.46 & 36.77  \\
        \textit{~w/o DI}  & 57.90 & 62.50 & 28.52 & 29.78 & 35.64 & 39.99 & 42.39  \\
        \textit{~w/o RCT} & 58.37 & 62.78 & 30.31 & 30.87 & 37.23 & \textbf{43.73} & 43.88  \\
        \textit{~w/o FDR} & 57.03 & 60.23 & 29.27 & 29.85 & 35.59 & 39.39 & 41.89  \\
        \textit{~w/o RG}  & 52.27 & 62.09 & 27.52 & 29.33 & 35.14 & 38.38 & 40.79  \\

    \bottomrule
    \end{tabular}
    }
    \caption{The ablation results (micro-F1)  averaged over three backbones. The best results are in \textbf{bold}.}
    \label{tab:ablation}
\end{table}

\subsection{Analysis}

\paragraph{Effectiveness on Various-scale LLMs}
To examine whether the proposed method remains effective for various-scale LLMs, we conduct experiments on various sizes of LLMs from the \textit{Llama-2-chat}~\cite{touvron-2023-llama2}, \textit{Meta-Llama-3-Instruct}~\cite{meta-2023-llama3}, and \textit{GPT-3.5}~\cite{rlhf-2022-chatgpt}, and present their results in Table~\ref{tab:main_results_2}.

From Table~\ref{tab:main_results_2}, it can be seen that our rationale supervisor can boost the performance of LLMs with various sizes.
Specifically, even with the most powerful Meta-Llama-3-70B-Instruct, there is still a 2.47\% micro-F1 score improvement over the original in-context learning.
The experimental results indicate that the ``relation bias'' issue exists in LLMs of various scales, and our proposed method can function as a plug-in module for various LLMs to effectively mitigate this problem.\looseness-1

\begin{table}[t!]
\small
    \centering
    \setlength{\tabcolsep}{0.5mm}
    {
    \begin{tabular}{lccccccc}
    \toprule
        \multirow{2}{*}{\textbf{Method}}  & \multicolumn{2}{c}{\textbf{SemEval}} & \multicolumn{2}{c}{\textbf{TACRED}}& \multicolumn{2}{c}{\textbf{Re-TACRED}} & \multirow{2}{*}{\textbf{Avg.}}\\
        \cmidrule(lr){2-3} \cmidrule(lr){4-5}\cmidrule(lr){6-7}
        & {{5-shot}} & {{10-shot}}  & {{5-shot}} & {{10-shot}} & {{5-shot}} & {{10-shot}}\\
        \midrule
        {R-BERT}     & 42.75&57.25&9.87&16.24&26.64&35.01&31.29 \\
        {KnowPrompt} & 53.92&56.42&27.86&30.34&50.08&55.41&45.67 \\
        \midrule
        \multicolumn{8}{c}{\textit{Llama-2-7b-chat}} \\
        \midrule
        {ICL}   & 58.68&64.90&26.11&26.35&42.75&45.53&44.05 \\
        {~w/ SRVF}          & 62.12&67.03&30.50&30.92&46.13&48.09&47.47\\
        \midrule
        \multicolumn{8}{c}{\textit{Llama-2-70b-chat}} \\
        \midrule
        {ICL}  & 68.92 & 69.86 & 27.32 & 27.12 & 43.63 & 44.94 & 46.97 \\
        {~w/ SRVF}         & 69.97 & 70.00 & 27.69 & 29.47 & 45.13 & 46.93 & 48.20 \\
        \midrule
        \multicolumn{8}{c}{\textit{Meta-Llama-3-8B-Instruct}} \\
        \midrule
        {ICL}  & 69.90 & 69.79 & 32.63 & 32.26 & 48.23 & 50.69 & 50.58 \\
        {~w/ SRVF}         & 71.14 & 71.41 & 35.26 & 34.29 & 52.23 & 55.25 & 53.26 \\
        \midrule
        \multicolumn{8}{c}{\textit{Meta-Llama-3-70B-Instruct}} \\
        \midrule
        {ICL}  & 71.21 & 72.40 & 34.71 & 34.97 & 56.10 & 57.41 & 54.47 \\
        {~w/ SRVF}         & 74.68 & 74.33 & 37.05 & 36.35 & 59.27 & 59.96 & 56.94\\
        \midrule
        \multicolumn{8}{c}{\textit{GPT-3.5-turbo}} \\
        \midrule
        {ICL}  & 67.26 & 70.58 & 32.46 & 31.38 & 43.56 & 46.88 & 48.69 \\
        {~w/ SRVF}         & 69.62 & 71.67 & 37.78 & 34.63 & 46.22 & 49.66 & 51.60  \\
    \bottomrule
    \end{tabular}
    }
    \caption{Results (micro-F1 scores) using various LLMs with the task-specific retriever.}
    \label{tab:main_results_2}
\end{table}

\paragraph{Comparision with Well-designed Few-shot Methods for RE}
As shown in Table~\ref{tab:main_results_2}, we include two established supervised fine-tuning methods for RE as baselines: 1) R-BERT~\cite{wu-2019-rbert}, which fine-tunes a BERT for the RE task, and 2) KnowPrompt~\cite{chen-2022-knowprompt}, which is tailored for few-shot scenarios and has shown good few-shot performance.
As we can see from the results, with the help of our proposed SRVF, even the relatively weak \textit{Llama-2-7b-chat} can outperform KnowPrompt by 1.80\% averagely.
Moreover, when deploying our SRVF on the most powerful \textit{Meta-Llama-3-70B-Instruct}, there is an average performance improvement of 11.27\% compared to KnowPrompt.

\paragraph{Analysis on Successfully Corrected Samples}

\begin{figure}[t]
\includegraphics[width=1.0\linewidth]{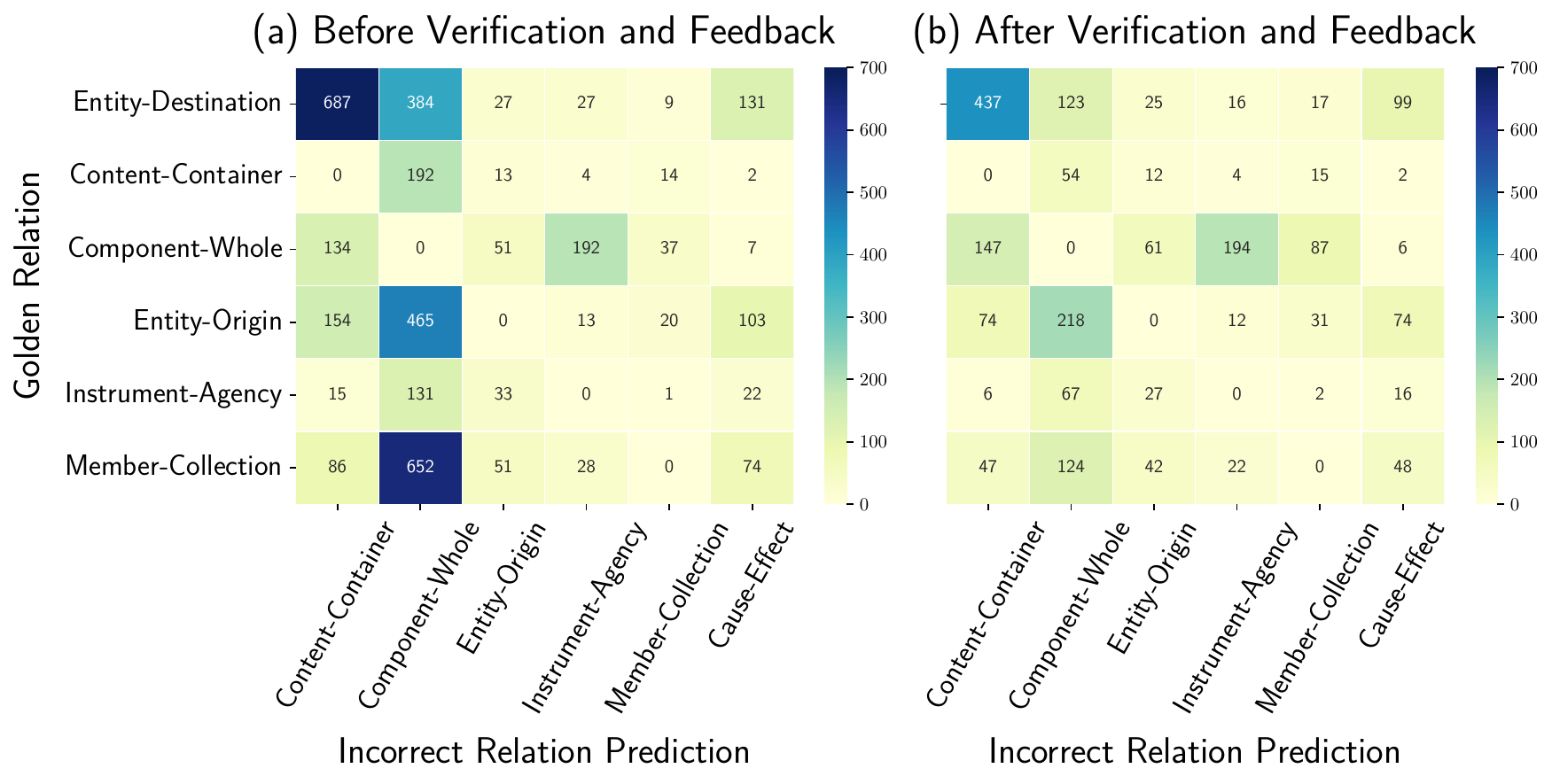}
\caption{Error matrix before and after the verification-feedback-correction procedure. The numbers show how many samples labeled $y$ (on the vertical axis) are incorrectly predicted as $x$ (on the horizontal axis).}
\label{fig:correction_matrix}
\end{figure}

To visualize which samples are successfully corrected by the proposed method, we compare the error matrix on the SemEval dataset before and after correction. The results are obtained by summing the number of error predictions of all settings in Table~\ref{tab:main_results_1}.
The results are shown in Fig.~\ref{fig:correction_matrix}.

From Fig.~\ref{fig:correction_matrix} (a), we observe that LLMs struggle to distinguish between relations that share similar entities, e.g., 687 samples labeled as ``Entity-Destination" are incorrectly predicted as ``Content-Container". 
Such error can arise when, for example, given sentences ``please move the eggs into the box" and ``there are 5 eggs in the box", where the same entity pair ``eggs" and ``box" form ``Entity-Destination" and ``Content-Container" relations, respectively. 
Such ambiguity often leads LLMs to misclassify relations when they fail to focus on context, resulting in numerous errors.
However, as shown in Fig.~\ref{fig:correction_matrix} (b), the number of samples labeled as ``Entity-Destination" but incorrectly predicted as ``Content-Container" is reduced by 250. This indicates that our method effectively alleviates the above issue.

\begin{figure}[t]
\centering
\includegraphics[width=0.7\linewidth]{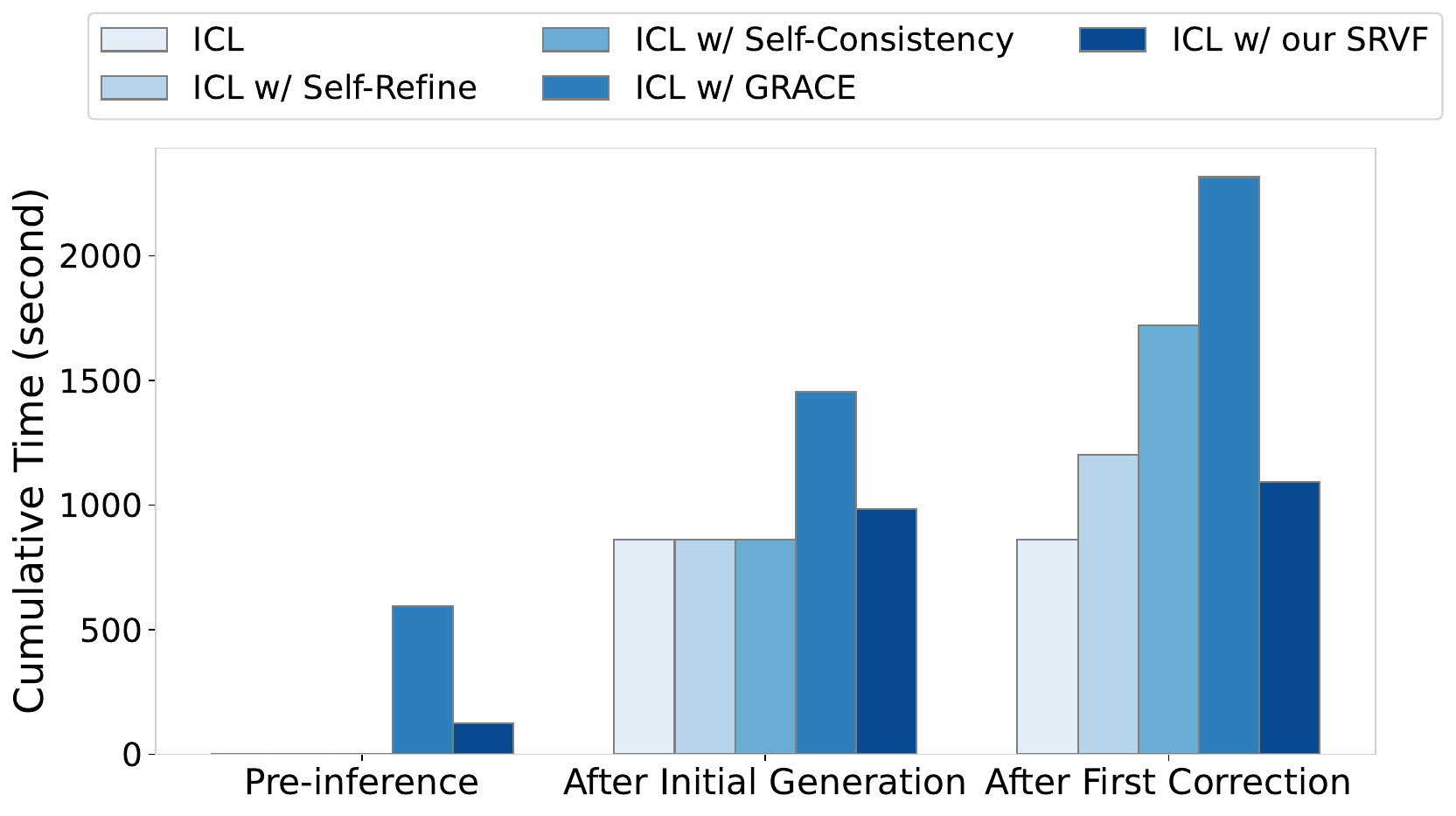}
\caption{Efficiency comparison of different methods on the 5-shot SemEval setting.
The results are accumulated along the X axis. For example, ``After Initial Generation" refers to the sum time of ``pre-inference'' and ``initial generation''.
}
\label{fig:inference_efficiency}
\end{figure}

\paragraph{Analysis on Method Efficiency}
Considering possible concerns on the inference efficiency due to the iterative feedbacks, we compare the inference time on the SemEval dataset of different methods.
Besides, we also evaluate the pre-inference time of each method, e.g., the time to obtain biased/unbiased data and train the rationale supervisor in our SRVF.
The comparison results are shown in Fig.~\ref{fig:inference_efficiency}.

From Fig.~\ref{fig:inference_efficiency}, we can observe that:

1) Basic in-context learning (ICL) is the most efficient.

2) Self-Refine does not require pre-inference time, but its inference time is more than the sum of our pre-inference time and inference time. Moreover, Self-Refine has the worst performance among all methods (Table~\ref{tab:main_results_1}).

3) Self-Consistency and GRACE have much higher computational costs than our SRVF, especially in terms of inference time.
This is mainly because the proposed rationale supervisor can verify whether the LLM prediction is biased. Only the test samples verified as biased by the rationale supervisor will proceed to the correction round for regeneration.
This greatly reduces the time cost of our method in inference time after correction.

Overall, our SRVF is the second-best in computational efficiency while achieving the best performance (Table~\ref{tab:main_results_1}).

\paragraph{Experiments on Document-level RE}\label{sec:doc_RE}

To explore the effectiveness of our method for document-level RE, we apply SRVF on three backbones and conduct experiments on two commonly used document-level RE datasets, DocRED~\cite{yao-2019-docred} and Re-DocRED~\cite{tan-2022-redocred}.
The random and SimCSE backbones are kept the same as before.
For the task-specific backbone, we borrow the idea from REPLM~\cite{ozyurt-2023-icl-llms-docre}, which obtains the final prediction by aggregating the predictions based on multiple retrieved demonstrations.
The experimental results are reported in Table~\ref{tab:doc_re_results}.

\begin{table}[t!]
\small
    \centering
    \setlength{\tabcolsep}{0.5mm}
    {
    \begin{tabular}{lccccc}
    \toprule
        \multirow{2}{*}{\textbf{Method}}  & \multicolumn{2}{c}{\textbf{DocRED}} & \multicolumn{2}{c}{\textbf{Re-DocRED}}& \multirow{2}{*}{\textbf{Avg.}}\\
        \cmidrule(lr){2-3} \cmidrule(lr){4-5}
        & {{5-shot}} & {{10-shot}}  & {{5-shot}} & {{10-shot}}\\
        \midrule
        {ICL (Random)}   & 7.76 & 7.82 & 8.27 & 7.73 & 7.90 \\
        {~w/ our SRVF}   & \textbf{15.40} & \textbf{18.00} & \textbf{15.29} & \textbf{15.65} & \textbf{16.09} \\
        \midrule
        {ICL (SimCSE)}   & 15.67 & 16.40 & 11.97 & 12.53 & 14.14 \\
        {~w/ our SRVF}   & \textbf{18.39} & \textbf{21.55} & \textbf{17.38} & \textbf{17.87} & \textbf{18.80} \\
        \midrule
        {ICL (Task-specific)}  & 18.29 & 18.40 & 17.44 & 18.67 & 18.20 \\
        {~w/ our SRVF}         & \textbf{20.04} & \textbf{21.55} & \textbf{19.98} & \textbf{21.69} & \textbf{20.82} \\
    \bottomrule
    \end{tabular}
    }
    \caption{Results (micro-F1) on the DocRED (document-level RE task). The best results are in \textbf{bold}.}
    \label{tab:doc_re_results}
\end{table}

From Table~\ref{tab:doc_re_results}, we can observe that: 
1) LLM performs poorly on document-level RE, which is consistent with empirical observations in ~\citet{li-2023-semi-llms-docre,sun-2024-consistency-llms-docre}. 
This is due to the difficulty LLMs face in selecting entity pairs that have certain relations from a vast space of candidate entity pairs.
Besides, the large number of candidate relation labels (96 in DocRED and Re-DocRED) further increases the difficulty in assigning each entity pair a relation.
2) Our proposed SRVF effectively enhances the performance of LLM under various settings on DocRED and Re-DocRED, indicating that our method remains to be effective in such challenging scenarios.

\section{Conclusion}
In this paper, we propose a novel automated feedback framework for LLM based relation extraction (RE), which includes a rationale supervisor to iteratively correct the biased relation prediction of LLMs.
Specifically, we first present a causal intervention and observation method to collect unbiased and biased rationales, which are then used to train the rationale supervisor.
Then, we develop a verification-feedback-correction procedure to iteratively enhance LLMs' ability to correct the biased prediction.
Extensive experiments demonstrate the superiority of our framework over existing methods.
In the future, we will try to extend the proposed framework to other NLP tasks~\footnote{We have done a preliminary study for the event detection task. Please refer to Appendix.}.

\section{Acknowledgments}
This work was supported by the grant from the National Natural Science Foundation of China (NSFC) project (No. 62276193).

\bibliography{aaai25}

\clearpage
\newpage

\section{Experimental Details}\label{sec:app:experimental_details}
\subsection{Datasets}
Table~\ref{tab:details_of_datasets_SenRE} shows the statistics of datasets used in our experiments and the number of training samples under various few-shot settings.

\begin{table}[htb]
    \centering
    \setlength{\tabcolsep}{3mm}
    \resizebox{\linewidth}{!}{
    \begin{tabular}{lcccc}
    \toprule
        \textbf{Dataset} & \textbf{Settings} & \textbf{\#Labels}  & \textbf{\#Train} & \textbf{\#Test} \\
        \midrule
        \multirow{4}{*}{SemEval} & 5-shot & \multirow{4}{*}{10}  & 50  &  \multirow{4}{*}{2717}  \\
         & 10-shot &   & 100 &    \\
         & 20-shot &   & 200 &    \\
         & 50-shot &   & 500 &    \\
        \midrule
        \multirow{4}{*}{TACRED} & 5-shot & \multirow{4}{*}{41}  & 210 &  \multirow{4}{*}{15433}  \\
         & 10-shot &   & 416 &    \\
         & 20-shot &   & 826 &    \\
         & 50-shot &   & 1994 &    \\
        \midrule
        \multirow{4}{*}{Re-TACRED} & 5-shot & \multirow{4}{*}{40}  & 200 &  \multirow{4}{*}{13373}  \\
         & 10-shot &   & 396 &    \\
         & 20-shot &   & 786 &    \\
         & 50-shot &   & 1898 &    \\
    \bottomrule
    \end{tabular}
    }
    \caption{Statistics of datasets used in our experiments.}
    \label{tab:details_of_datasets_SenRE}
\end{table}

\subsection{Backbones}
In this paper, we focus on relation extraction using LLMs.
This section will show the details of basic prompt structure design and demonstration selection strategies that are used as backbones.

\subsubsection{Prompt Design for LLM based RE}\label{sec:app:basic_prompt_llm_re}
Table~\ref{tab:prompt_for_llms_based_RE} presents an example of the designed prompt with the corresponding expected response.
The prompt consists of four parts, i.e., \{\texttt{Instruction}, \texttt{Demonstrations}, \texttt{Hint}, \texttt{Inference}\}.
When conducting relation extraction using LLMs, we input the prompt into LLMs and parse the response for the predicted relation.
Besides, due to the limited context size of LLMs~\footnote{For LLMs used in our experiments, i.e., Llama-2-chat family and \texttt{gpt-3.5-turbo-0613}, the maximum context size is 4096 (tokens).}, we set the number of demonstrations as 10, 4, and 4 for the SemEval, TACRED, and Re-TACRED datasets, respectively.

\subsubsection{Demonstration Selection Strategy}
\paragraph{Random} For the ``Random'' strategy, we randomly select $m$ in-context demonstrations from the few-shot labeled sample set.
\paragraph{SimCSE} For the ``SimCSE'' strategy, we retrieve the top-k nearest samples as in-context demonstrations measured by the sentence embedding. 
Specifically, we adopt the embeddings corresponding to the [CLS] tokens for retrieval.
In our experiments, we adopt the \texttt{sup-simcse-bert-base-uncased}~\cite{gao-2021-simcse} as the encoder to obtain the embeddings.
\paragraph{Task-specific} For the ``Task-specific'' strategy, similar to the ``SimCSE'' strategy, we retrieve the top-k nearest samples as in-context demonstrations based on certain embeddings. Different from the ``SimCSE'' strategy, here we adopt a task-specific encoder to obtain the task-aware embeddings, which is proposed by ~\citet{wan-2023-gpt_re} and is the state-of-the-art strategy to select in-context demonstrations for LLM based RE.
Specifically, we first concat each sample with special tokens, i.e., \{[CLS], \textit{sentence-part1}, \$, \textit{head-entity}, \textit{sentence-part2}, \#, \textit{tail-entity}, \textit{sentence-part3}, [SEP]\}.
We average the embeddings corresponding to [CLS], \textit{head-entity} and \textit{tail-entity} tokens as the embedding for retrieval.
The embeddings are obtained by a task-specific encoder, which is initialized as BERT~\cite{devlin-2018-bert} and is trained via the cross-entropy loss for classification using the few-shot labeled sample set.
For the training of the task-specific encoder, the batch size is set to 16, the learning rate is set to 2e-5, and the epoch number is set to 100.

\begin{table*}[t]
\centering
\setlength{\tabcolsep}{3mm}
\resizebox{\linewidth}{!}{
	\begin{tabular}{p{21cm}}
	\toprule
        \textit{Prompt of LLM based RE}\\
        \midrule
        \textbf{Instruction}: Determine the relation between the given head entity and tail entity in the given sentence. The relation category is from the relation type set.\\ 
        \textbf{Demonstrations}:\\ 
        Demo Index: 0 \\ 
        (Start of Instance)\\ 
        Given Sentence: "A maker of electric cars plans to use old GM plant in Delaware ."\\ 
        Relation Type Set: \{Other, Component-Whole, Instrument-Agency, Member-Collection, Cause-Effect, Entity-Destination, Content-Container, Message-Topic, Product-Producer, Entity-Origin\}\\ 
        Head Entity: "cars"\\ 
        Tail Entity: "maker"\\ 
        Reasoning Explanations: In the given sentence, the key phrase "maker of electric cars" implies that the maker is responsible for producing the electric cars. Therefore, the head entity "cars" serves as the "Product" while the tail entity "maker" serves as the "Producer". \\ 
        Prediction: Given the sentence, the relation between the head entity "cars" and the tail entity "maker" is "Product-Producer". \\ 
        (End of Instance)\\ 
         Please learn the demonstration and follow the instruction, complete the "Reasoning Explanations" and "Prediction" parts of the new given instance. You only need to solve the only instance given. Please end with (End of Instance) when complete the text. \\ 
        \textbf{Inference}:\\ 
        (Start of Instance)\\ 
        Given Sentence: "I 'm going with some girls to get a bunch of flowers ."\\ 
        Relation Type Set: \{Other, Component-Whole, Instrument-Agency, Member-Collection, Cause-Effect, Entity-Destination, Content-Container, Message-Topic, Product-Producer, Entity-Origin\} \\ 
        Head Entity: "flowers" \\ 
        Tail Entity: "bunch" \\ 
        \midrule        
        \textit{Response of LLM based RE}\\
        \midrule        
        Reasoning Explanations: In the given sentence, the key phrase "bunch of flowers" implies that the flowers are the individual elements that make up the collection of flowers. Therefore, the head entity "flowers" serves as the "Member" while the tail entity "bunch" serves as the "Collection".  \\ 
        Prediction: Given the sentence, the relation between the head entity "flowers" and the tail entity "bunch" is "Member-Collection". \\ 
        (End of Instance)\\
	\bottomrule
	\end{tabular}
 }
 \caption{An example of the designed prompt with the corresponding expected response. The example is from the SemEval dataset. Note there is only one demonstration included due to the space limit.}
 \label{tab:prompt_for_llms_based_RE}

\end{table*}

\begin{table*}[t]
\centering
\setlength{\tabcolsep}{3mm}
\resizebox{\linewidth}{!}{
	\begin{tabular}{p{21cm}}
	\toprule
        \textit{Prompt for verification of Self-Refine}\\
        \midrule
        \textbf{Instruction}: Check for possible types of errors such as inconsistency in the following prediction results and rationale for judgments about a relationship, and give reasons for the judgments.\\ 
        \textbf{Demonstrations}:\\ 
        (Start of Instance)\\ 
        Given Sentence: "Space X was founded by Musk."\\ 
        Head Entity: "Space X"\\ 
        Tail Entity: "Musk"\\ 
        Reasoning Explanations: In the given sentence, the key phrase "was founded by" implies that the company "Space X" was created by the person "Musk". Therefore, the head entity "Space X" serves as the "org" while the tail entity "Musk" servers as the "founded\_by" person. \\ 
        Prediction: The relation type between "Space X" and "Musk" is "org:founded\_by"\\ 
        Check Results: There's no error here.\\ 
        (End of Instance)\\ 
        (Start of Instance)\\ 
        Given Sentence: "Space X was founded by Musk."\\ 
        Head Entity: "Space X"\\ 
        Tail Entity: "Musk"\\ 
        Reasoning Explanations: In the given sentence, the key phrase "was founded by" implies that the company "Space X" was created by the person "Musk". Therefore, the head entity "Space X" serves as the "org" while the tail entity "Musk" serves as the "top\_members/employees" person. \\ 
        Prediction: The relation type between "Space X" and "Musk" is "org:top\_members/employees"\\ 
        Check Results: There's an inconsistency error here. As in the explanation given, the specific reasoning process therein is correct for the roles of Musk and SpaceX in the relationship, but ultimately there is an inconsistency issue when the answer is given.\\ 
        (End of Instance)\\ 
         \\
        Please learn the demonstration and follow the instruction, output the explanations part of the new given instance. \\
        Please end with (End of Instance) when completing the text.\\
        \\ 
        \textbf{Inference}:\\ 
        (Start of Instance)\\ 
        Given Sentence: "\{given\_sentence\}"\\ 
        Head Entity: "\{head\_entity\}"\\ 
        Tail Entity: "\{tail\_entity\}"\\ 
        Reasoning Explanations: \{pred\_explanations\}\\ 
        Prediction: The relation type between "\{head\_entity\}" and "\{tail\_entity\}" is "\{pred\_label\}"\\ 
        Check Results: \\ 
        \midrule        
        \textit{Prompt for refinement of Self-Refine}\\
        \midrule   
        \textbf{Special Instruction}: Please take care to avoid the following mistakes when making predictions.\\ 
        \textbf{Potential Mistakes}: \{Error feedback provided at the verification stage\}\\
        \{Initial prompt for LLM based RE\}\\
	\bottomrule
	\end{tabular}
 }
 \caption{An example of the designed prompt for the verification and refinement of the reproduced Self-Refine. Note the \{Initial prompt for LLM based RE\} denotes the prompt used in Appendix~\ref{sec:app:basic_prompt_llm_re} because without it the performance will degrade to worse zero-shot RE.}
 \label{tab:prompt_for_self_refine_RE}
\end{table*}

\subsection{Baselines}\label{sec:app:baselines_prompts_feedback_methods}
\paragraph{Self-Refine}~\cite{madaan-2023-self_refine} is proposed to improve the initial response from LLMs through iterative feedback and refinement.
Since it is designed for various reasoning tasks, e.g., math reasoning, code optimization, or acronym generation, we cannot directly adopt it for the RE task.
Thus, to adapt it for RE, we design the ``inconsistency issue'' as the verification objective, which means that the given rationale is not consistent with the final relation prediction.
Specifically, this framework consists of three LLM based agents, i.e., RE agent, verifier agent, and refiner agent.
First, we prompt the RE agent using the designed prompt in Appendix~\ref{sec:app:basic_prompt_llm_re} for relation extraction.
Second, we ask the verifier agent to find errors like the ``inconsistency issue'' in the initial response.
If the verifier agent finds certain errors, it will output the corresponding reasons for the found error, which will be treated as feedback for refinement.
Finally, the error feedback given by the verifier agent will be fed to the refiner agent to correct the initial prediction.
To show how this approach is reproduced, we show an example of the prompt used for the verification and refinement in Table~\ref{tab:prompt_for_self_refine_RE}. 

\paragraph{Self-Consistency}~\cite{wang-2023-selfconsistency} is proposed to conduct verification for the multiple candidate responses.
Specifically, it employs a simple majority voting strategy to select the final prediction from the candidate predictions.
For example, if there are 5 candidate relation predictions $\{y_A, y_B, y_A, y_A, y_A\}$, the final prediction is $y_A$, which is the majority one.

\paragraph{GRACE}~\cite{khalifa-2023-grace} is proposed to conduct verification and feedback for multi-step reasoning tasks.
It selects the best one for each intermediate reasoning step, where the selected reasoning step is used as the feedback for LLMs to generate the next step.
However, the reasoning procedure (rationale) in the RE task usually consists of only one step, making such feedback unsuitable for RE.
Thus, in the experiments, we only use it for the verification stage, i.e., selecting the best relation prediction via selecting the corresponding best rationale.
Specifically, we follow its official code to implement the training of the discriminator, which is used for calculating the score for each candidate rationale in the inference time.


\subsection{Implement Details}
\subsubsection{Implementation of LLMs}
\paragraph{Open-sourced LLMs}
For the \textit{Llama-2-chat} and \textit{Meta-Llama-3-Instruct} family LLMs, to improve the generation efficiency of LLMs, we adopt the accelerated inference framework vLLM~\cite{kwon-2023-vllm}, which utilizes \textit{PagedAttention} to increase the speed of LLMs to process prompts and output responses in batches.
We use NVIDIA A800 80G as the experimental equipment.

\paragraph{Closed-sourced LLMs}
We use \texttt{gpt-3.5-turbo-0613} API from OpenAI to implement the GPT-3.5-turbo.
For experiments of the proposed method, the temperature parameter, which is used to control the output diversity, is set to 1.
For the baselines, i.e., Self-Refine, Self-Consistency, and GRACE, the temperature is set as 0.7 to increase the diversity of generated responses.

\subsubsection{Implementation of the Proposed SRVF}\label{sec:app:implementation_of_method}
In this section, we will introduce the implementation details of the proposed automated feedback method, which is described in Section~\ref{sec:method}.

\paragraph{Causal Intervention and Observation}
The proposed causal intervention and observation method is designed to collect unbiased and biased rationales, which are then used for training the rationale supervisor.
Specifically, the diversified intervention (DI) prompt used for observing biased rationales is similar to the LLM based RE prompt (Appendix~\ref{sec:app:basic_prompt_llm_re}).
The only difference is that the in-context demonstration here is set to a labeled sample with a different label than the observed labeled sample.
For the label-guided intervention (LGI) prompt used to induce unbiased rationales, we show an example in Table~\ref{tab:prompt_for_induce_unbiased_rationales}.
Besides, to explore the sensitivity of the two prompts used here, we present a prompt sensitivity analysis in Appendix~\ref{sec:app:analysis_of_prompt_sensitivity}.\looseness-1

\paragraph{Rationale Supervisor Training}
We adopt the commonly used pre-trained language model BERT~\cite{devlin-2018-bert} as the initialization of the rationale supervisor.
The temperature hyper-parameter $\tau$ used in Eq.~\ref{eq:rationale_contrastive_loss} is set to 0.2 for all experiments~\footnote{We explore the impact of the hyper-parameter $\tau$ in Appendix~\ref{sec:app:impact_of_tau_of_cl}.}.
For the training of the rationale supervisor, the batch size is set to 128, the learning rate is set to 2e-5, and the epoch number is set to 50.

\paragraph{Feedback Demonstration Retrieval}
We set the number of feedback demonstrations as 5, 4, and 4 for the SemEval, TACRED, and Re-TACRED datasets, respectively~\footnote{We explore the impact of the number of feedback demonstrations in Appendix~\ref{sec:app:impact_feedback_demo_num}.}.

\begin{table}[htp]
    \centering
    \setlength{\tabcolsep}{1mm}
    \resizebox{\linewidth}{!}{
    \begin{tabular}{lcccccc}
    \toprule
        \multirow{2}{*}{\textbf{Dataset}} & \multirow{2}{*}{\textbf{Settings}} & \multirow{2}{*}{\textbf{\#Labels}}  & \multicolumn{2}{c}{\textbf{\#Train}} & \multicolumn{2}{c}{\textbf{\#Test}} \\
        \cmidrule(lr){4-5}\cmidrule(lr){6-7}
         & & & \textit{{\#Docs}} & \textit{{\#Triplets}} & \textit{{\#Docs}} & \textit{{\#Triplets}} \\
        \midrule
        \multirow{2}{*}{DocRED} & 5-shot & \multirow{2}{*}{96}  & 38  & 481 &  \multirow{2}{*}{998} &  \multirow{2}{*}{12275}  \\
         & 10-shot &   & 79 & 972 &  &    \\
        \midrule
        \multirow{2}{*}{Re-DocRED} & 5-shot & \multirow{2}{*}{96}  & 19 & 497 &  \multirow{2}{*}{1000} &  \multirow{2}{*}{34732}  \\
         & 10-shot &   & 36 & 986 &  &    \\
    \bottomrule
    \end{tabular}
    }
    \caption{Statistics of datasets used in our document-level relation extraction experiments.}
    \label{tab:details_of_datasets_DocRE}
\end{table}

\begin{table*}[t]
\centering
\setlength{\tabcolsep}{3mm}
\resizebox{\linewidth}{!}{
	\begin{tabular}{p{20.5cm}}
	\toprule
        \textit{Prompt for inducing unbiased rationales (step 1)}\\
        \midrule
        \textbf{Instruction}: Given a sentence, explain why there is certain relation between the head and tail entities in the sentence. \\ 
        \textbf{Demonstrations}: \\ 
        (Start of Instance)\\ 
        Given Sentence: "The therapist treats the patient with a certain kind of manual therapy."\\ 
        Head Entity: "therapy"\\ 
        Tail Entity: "therapist"\\ 
        The relation type between "therapy" and "therapist" is "Instrument-Agency"\\ 
        Reasoning Explanations: In the given sentence, the key phrase "therapist treats the patient with a certain kind of manual therapy" implies that the therapy is the tool employed by the therapist to treat the patient. Therefore, the head entity "therapy" serves as the "Instrument" while the tail entity "therapist" servers as the "Agency".\\ 
        Prediction: Given the sentence, the relation between the head entity "therapy" and the tail entity "therapist" is "Instrument-Agency"\\ 
        (End of Instance) \\ 
         \\
        Please learn the demonstration and follow the instruction, complete the "Reasoning Explanations" and "Prediction" parts of the new given instance.\\
        Please end with (End of Instance) when complete the text. \\
        \\ 
        \textbf{Inference}:\\ 
        (Start of Instance)\\
        Given Sentence: "The fueltruck was contained in a large box to ensure that any spilled diesel would be contained ."\\ 
        Head Entity: "fueltruck"\\ 
        Tail Entity: "box"\\ 
        The relation type between "fueltruck" and "box" is "Content-Container"\\ 
        \midrule        
        \textit{Response for inducing unbiased rationales (step 1)}\\
        \midrule  
        Reasoning Explanations: In the given sentence, the key phrase "fueltruck was contained in a large box" implies that the fueltruck is the content being stored in the box. Therefore, the head entity "fueltruck" serves as the "Content" while the tail entity "box" serves as the "Container".\\ 
        Prediction: Given the sentence, the relation between the head entity "fueltruck" and the tail entity "box" is "Content-Container".\\ 
        (End of Instance)\\ 
        \midrule        
        \midrule        
        \textit{Prompt for inducing unbiased rationales (step 2)}\\
        \midrule   
        \textbf{Instruction}: Given a sentence and corresponding explanations, try to derive the relation label prediction.\\ 
        \textbf{Demonstrations}:\\ 
        (Start of Instance)\\
        Given Sentence: "The therapist treats the patient with a certain kind of manual therapy ."\\
        Relation Type Set: \{Other, Component-Whole, Instrument-Agency, Member-Collection, Cause-Effect, Entity-Destination, Content-Container, Message-Topic, Product-Producer, Entity-Origin\}\\
        Head Entity: "therapy"\\
        Tail Entity: "therapist"\\
        Reasoning Explanations: In the given sentence, the key phrase "therapist treats the patient with a certain kind of manual therapy" implies that the therapy is the tool employed by the therapist to treat the patient. Therefore, the head entity "therapy" serves as the "Instrument" while the tail entity "therapist" servers as the "Agency".\\
        Based on the above reasoning explanations, the relation between the head entity "therapy" and the tail entity "therapist" is "Instrument-Agency"\\
        (End of Instance)\\
         \\
        Please learn the demonstration and follow the instruction, output the inference result of the new given instance.\\
        \\ 
        \textbf{Inference}:\\ 
        (Start of Instance)\\
        Given Sentence: "The fueltruck was contained in a large box to ensure that any spilled diesel would be contained ."\\
        Relation Type Set: \{Other, Component-Whole, Instrument-Agency, Member-Collection, Cause-Effect, Entity-Destination, Content-Container, Message-Topic, Product-Producer, Entity-Origin\}\\
        Head Entity: "fueltruck"\\
        Tail Entity: "box"\\
        Reasoning Explanations:  In the given sentence, the key phrase "fueltruck was contained in a large box" implies that the fueltruck is the content being stored in the box. Therefore, the head entity "fueltruck" serves as the "Content" while the tail entity "box" serves as the "Container".\\
        Based on the above reasoning explanations, \\
        \midrule        
        \textit{Response for inducing unbiased rationales (step 2)}\\
        \midrule  
         the relation between the head entity "fueltruck" and the tail entity "box" is "Content-Container"\\
        (End of Instance)\\
	\bottomrule
	\end{tabular}
 }
 \caption{An example of the designed prompts for inducing unbiased rationales. Step 1 corresponds to obtaining the unbiased rationale based on the given golden label. Step 2 corresponds to the process of checking the consistency between the obtained unbiased rationale with the golden label.}
 \label{tab:prompt_for_induce_unbiased_rationales}
\end{table*}


\subsection{Experimental Details of SRVF for Document-level Relation Extraction}
This section will present details of experiments for document-level relation extraction (RE) in Section~\ref{sec:doc_RE}.

\paragraph{Datasets}
Table~\ref{tab:details_of_datasets_DocRE} shows the statistics of datasets used in our experiments for document-level RE, including the number of documents (\textit{\#Docs}) and triplets (\textit{\#Triplets})~\footnote{Since the labeled test set of DocRED is not publicly available, we use the original development set as the test set. Additionally, for the Re-DocRED dataset, we combine the original development set and the test set to form a unified test set.}.

\paragraph{Few-shot Settings}
To keep consistent with the main experimental setup of this paper, for the document-level RE, we also adopt the $k$-shot ( $k\in \{5,10\}$) settings. 
Specifically, we employ a greedy sampling method to select the $k$-shot training set. 
The sampling stops when the average number of triplets in the $k$-shot document sample set, calculated as the total number of triplets divided by the number of relation labels, exceeds $k$. 
As shown in Algorithm~\ref{alg:greedy_sampling_k_shot}, we continuously and randomly select candidate document samples $s$. If $s$ contains any triplet of any relation label that has not yet met its sampling quota ($k$-shot per relation label), we add $s$ to the $k$-shot training set; otherwise, we discard it and select another candidate document sample. This process repeats until the stopping criterion is met.

\paragraph{Backbones}
For document-level RE, we apply the proposed SRVF on three different backbones:
\begin{itemize}
    \item \textit{Random} randomly selects samples from the given labeled data as demonstrations for in-context learning.
    \item \textit{SimCSE} retrieves the top-k nearest samples from the given labeled data as demonstrations. Specifically, we use the embeddings corresponding to the [CLS] tokens of the document texts for retrieval. The embeddings are obtained using the \texttt{sup-simcse-bert-base-uncased}~\cite{gao-2021-simcse} encoder.
    \item \textit{Task-specific}~\cite{ozyurt-2023-icl-llms-docre} first retrieves $k$ stes of demonstrations and then uses in-context learning to predict $k$ sets of triplets. Finally, the $k$ sets are aggregated and selected to form the final prediction set.
\end{itemize}

\paragraph{Prompt Design}
Considering that the document-level RE task includes entity pair matching and relation prediction, following~\cite{wei-2023-zero_shot_IE_chatgpt}, we adopt a two-stage in-context learning approach for document-level RE.
Specifically, for the first stage, we aim to predict candidate entity pairs that potentially have certain relations, i.e., \{\texttt{Candidate Entity Pairs}\} = LLM(\{\texttt{Instruction}, \texttt{Demonstration}, \texttt{Hint}, \texttt{Test Document}, \texttt{Candidate Entities}\}).
For the second stage, we aim to predict relation triplets based on the candidate entity pairs, i.e., \{\texttt{Relation Triplets}\} = LLM(\{\texttt{Instruction}, \texttt{Demonstration}, \texttt{Hint}, \texttt{Test Document}, \texttt{Candidate Entity Pairs}\}).
Besides, due to the context window length limitation of LLMs, we set the number of in-context demonstrations as one~\footnote{Although recent long-context LLMs allow for more in-context demonstrations in the prompt, we observe in our experiments that this significantly increases inference time without noticeably enhancing performance.}. We present an example in Table~\ref{tab:basic_prompt_example_for_two_stage_docre} to help understand the above design.

\paragraph{Implementation Details}
For document-level RE, the core modules of the proposed SRVF, including causal intervention and observation, rationale supervisor training, and feedback demonstration retrieval, are kept the same as sentence-level RE. 
However, since document-level RE requires the prediction of multiple relation triplets for a single document sample, we adaptively incorporate a strategy to retain or discard predicted relation triplets after the verification~\footnote{The sentence-level RE requires only one relation prediction per sample and thus does not need this strategy)}. This strategy is presented as Algorithm~\ref{alg:prediction_process_for_DocRE}.
For LLM, we adopt the Meta-Llama-3-8B-Instruct considering its long context window (8196 tokens). 
Besides, the number of feedback iterations is set to 5 for all settings.

\begin{algorithm}
\begin{algorithmic}[1]
\REQUIRE $k$, $D$ (original full training samples set), $R$ (relation labels set)
\ENSURE $S_k$ ($k$-shot training set)
\STATE Initialize $S_k \gets \emptyset$
\STATE Initialize $q \gets 0$
\WHILE{$q \leq k$}
    \STATE Randomly select $s \in D$
    \IF{$s$ contains triplets of unfinished relation types}
        \STATE $S_k \gets S_k \cup \{s\}$
    \ENDIF
    \STATE Calculate $q \gets \frac{|\text{Total triplets in } S_k|}{|R|}$
\ENDWHILE
\caption{Greedy Sampling for $k$-shot Setting of Document-level RE}
\label{alg:greedy_sampling_k_shot}
\end{algorithmic}
\end{algorithm}

\begin{algorithm}
\begin{algorithmic}[1]
\REQUIRE Sample $s$ (with document text and candidate entity set), relation labels set $R$, feedback iterations $m$
\ENSURE Predicted triplet set $T_{unbiased}$
\STATE Initialize $T_{unbiased} \gets \emptyset$
\STATE Randomly select or retrieve initial in-context demonstration $d$
\STATE $T \gets \text{LLM\_predict}(d, s, R)$
\FOR{iteration $i \gets 1$ \TO $m$}
    \STATE Initialize $T_{biased} \gets \emptyset$
    \FOR{each triplet $t \in T$}
        \IF{Rationale for $t$ is verified as unbiased by SRVF}
            \IF{$t \notin T_{unbiased}$}
                \STATE $T_{unbiased} \gets T_{unbiased} \cup \{t\}$
            \ENDIF
        \ELSE
            \STATE $T_{biased} \gets T_{biased} \cup \{t\}$
        \ENDIF
    \ENDFOR
    \STATE Select new in-context demonstration $d$ based on $T_{biased}$
    \STATE $T \gets \text{LLM\_predict}(d, s, R)$
\ENDFOR
\caption{Prediction Process of SRVF for Document-level RE}
\label{alg:prediction_process_for_DocRE}
\end{algorithmic}
\end{algorithm}

\begin{table*}[t]
\centering
\setlength{\tabcolsep}{3mm}
\resizebox{\linewidth}{!}{
	\begin{tabular}{p{20.5cm}}
	\toprule
        \textit{Prompt for extracting candidate entity pairs (stage 1)}\\
        \midrule
        (Instruction) Check the document, and find all the possible entity pairs that may hold certain relations. (/Instruction)\\
        (Demonstrations)\\
        (Instance)\\
        Given Document: "Berthe Marie Pauline Morisot ..."\\
        Candidate Relation Types: \{employer, capital, ... \}\\
        Candidate Entities: \{Berthe Marie Pauline Morisot, January 14, 1841, ... \}\\
        Candidate Entity Pairs: \\
        1. (Pair)(head)Berthe Marie Pauline Morisot(/head)(tail)January 14, 1841(/tail)(/Pair)\\
        ......\\
        (/Instance)\\
        (/Demonstrations)\\
        (Test)\\
        (Hint)The head and tail entity must be chosen from the Candidate Entities.(/Hint)\\
        (Instance)\\
        Given Document: Skai TV is a Greek free... .\\
        Candidate Relation Types: \{employer, capital, ... \}\\
        Candidate Entities: \{Skai TV, Greek, Piraeus, ... , Greece\}\\
        Candidate Entity Pairs: \\
        \midrule        
        \textit{Response for extracting candidate entity pairs (stage 1)}\\
        \midrule        
        1. (Pair)(head)Skai TV(/head)(tail)Piraeus(/tail)(/Pair)\\
        ...\\
        (/Instance)\\

        \midrule        
        \midrule        

        \textit{Prompt for predicting relations of candidate entity pairs (stage 2)}\\
        \midrule  
        (Instruction) Considering the document, and generate a triplet with a proper relation for each entity pair. The number of triplets must match the given entity pairs.(/Instruction)\\
        (Demonstrations)\\
        (Instance)\\
        Given Document: "Berthe Marie Pauline Morisot ... "\\
        Candidate Relation Types: \{employer, capital, ... \}\\
        Candidate Entity Pairs: 1. (Triplet)(head)Berthe Marie Pauline Morisot(/head)(tail)January 14, 1841(/tail)(/Triplet)\\
        ......\\
        Extracted Triplets: \\
        1. (Triplet)(head)Berthe Marie Pauline Morisot(/head)(relation)date of birth(/relation)(tail)January 14, 1841(/tail)(explanation) The key phrase "January 14, 1841 – March 2, 1895" implies that "January 14, 1841" is the "date of birth" of "Berthe Marie Pauline Morisot"... (/explanation)(/Triplet)\\
        ......\\
        (/Instance)\\
        (/Demonstrations)\\
        (Test)\\
        (Hint) The relation must be chosen from the given Candidate Relation Types. Please generate \{number of candidate pairs\} triplets that correspond exactly to the given entity pairs. (/Hint)\\
        (Instance)\\
        Given Document: "Skai TV is a Greek free - ..." \\
        Candidate Relation Types: \{employer, capital, ... \}\\
        Candidate Entity Pairs: \\
        1. (Triplet)(head)Skai TV(/head)(tail)Piraeus(/tail)(/Triplet)\\
        ......\\
        Extracted Triplets: \\

        \midrule  
        \textit{Response for predicting relations of candidate entity pairs (stage 2)}\\
        \midrule 
        1. (Triplet)(head)Skai TV(/head)(relation)headquarters location(/relation)(tail)Piraeus(/tail)(explanation) The sentence "Skai TV is a Greek free-to-air television network based in Piraeus" explicitly states that "Skai TV" is located in "Piraeus".(/explanation)(/Triplet)\\
        ......\\
        (/Instance)\\

	\bottomrule
	\end{tabular}
 }
 \caption{An example of the used basic prompt for the document-level RE task. Here, we have abbreviated some of the text due to space constraints.}
 \label{tab:basic_prompt_example_for_two_stage_docre}
\end{table*}

\clearpage
\newpage

\section{Supplementary Analysis}\label{sec:app:supplementary_experiments}

\subsection{Event Detection Task}\label{sec:app:framework_for_event_detection_task}

To explore the extensibility of our proposed approach to other NLP tasks, we select another important task in information extraction, i.e., event detection.
The event detection (ED) task aims to detect event trigger words in a given sentence and categorize the event trigger words. 
For example, given the sentence: \textit{``Shaunie O'Neal gave birth to the couple 's third child at 1:52.''}, we aim to extract that \textit{``\{birth\} is the trigger word for a \{Life:Be-Born\} event''}.
We select the widely used dataset ACE05~\cite{LDC-2005-ACE05_ED} for a preliminary study, which has 33 event types.
The experimental results are shown in Table~\ref{tab:event_detection_task}.

\begin{table}[htp!]
    \centering
    \setlength{\tabcolsep}{2mm}
    \resizebox{\linewidth}{!}{
    \begin{tabular}{lccccc}
    \toprule
        \multirow{2}{*}{\textbf{Method}}  & \multicolumn{4}{c}{\textbf{ACE05}} & \multirow{2}{*}{\textbf{Avg.}}\\
        \cmidrule(lr){2-5}
        & {{5-shot}} & {{10-shot}}  & {{20-shot}} & {{50-shot}} \\
        \midrule
        \multicolumn{6}{c}{\textit{Random}} \\
        \midrule
        In-context Learning & 12.26 & 14.63 & 14.52 & 12.72 & 13.53\\
        \textbf{~w/ our SRVF} &  \textbf{18.81} & \textbf{23.09} & \textbf{20.21} & \textbf{22.19} & \textbf{21.08}\\
        
        \midrule
        \multicolumn{6}{c}{\textit{SimCSE}} \\
        \midrule
        
        In-context Learning & 18.69 & 18.10 & 19.62 & 18.70 & 18.78\\
        \textbf{~w/ our SRVF} & \textbf{24.29} & \textbf{26.28} & \textbf{27.85} & \textbf{29.09} & \textbf{26.88} \\
        
        \midrule
        \multicolumn{6}{c}{\textit{Task-specific}} \\
        \midrule
        
        In-context Learning & 18.98 & 14.54 & 16.31 & 18.23 & 17.02\\
        \textbf{~w/ our SRVF} & \textbf{23.97} & \textbf{20.13} & \textbf{21.88} & \textbf{25.60} & \textbf{22.90} \\
        
    \bottomrule
    \end{tabular}
    }
    \caption{Results (micro-F1 scores) on the ACE05 dataset are reported using Llama-2-7b-chat with the three kinds of backbones. The best results are in \textbf{bold}.}
    \label{tab:event_detection_task}
    \vspace{-1mm}
\end{table}

As shown in Table~\ref{tab:event_detection_task}, we can observe that: 
1) The original in-context learning performance is very poor, which may be due to the weak ability of Llama-2-7b-chat to discover trigger words and categorize trigger words.
Besides, since the task-specific retriever proposed by~\citet{wan-2023-gpt_re} is designed for the relation extraction task, it cannot be well applied to the event detection task.
Thus, it is hard to improve the ED performance as the number of labeled samples increases.
2) The proposed SRVF method can boost the performance for all backbones under all settings, which indicates that the proposed method also works for the event detection task.

\subsection{Supplementary Analysis of SRVF}\label{sec:app:temp_supplementary_analysis_SRVF}

\begin{figure}[htbp]
\centering
\includegraphics[width=0.85\linewidth]{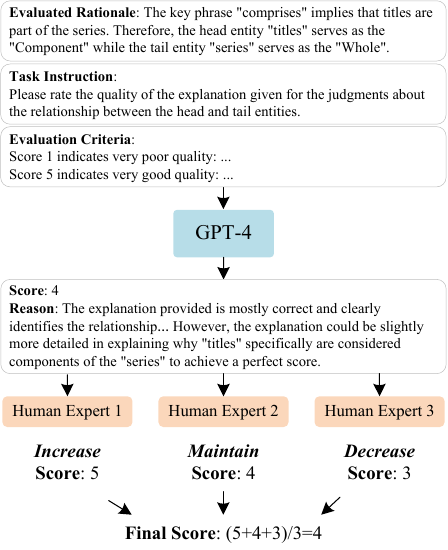}
\caption{Illustration of the quality evaluation procedure for the generated unbiased rationale.}
\label{fig:quality_analysis_method}
\end{figure}

\paragraph{Quality Analysis of Unbiased Rationale}
We conduct a quality analysis of the unbiased rationale generated by our proposed SRVF.
Specifically, the set for evaluation is composed of samples from SemEval, TACRED, and Re-TACRED under the 5-shot setting, with 50, 210, and 200 samples respectively.
Following~\citet{liu-2023-gpt4eval}, we first input the rationale, task introduction, and evaluation criteria to GPT-4, and ask it to generate a quality score with detailed reason.
Then, we ask three human experts to raise or lower the initial score when they disagree with the reason given by GPT-4.
The agreement rate between GPT-4 and the human experts is 91.3\%.
As shown in Fig.~\ref{fig:quality_analysis_method}, the overall process for evaluating the quality of the generated unbiased rationale is as follows:

\begin{itemize}
    \item \textbf{Step 1: Initial Scoring} As shown in Table~\ref{tab:quality_analysis_prompt_for_gpt4_evaluation}, we input the evaluated rationale, task instruction, and evaluation criteria to GPT-4~\footnote{Specifically, we adopt the \texttt{gpt-4o} API from OpenAI.} to score the rationale on a scale from 1 to 5. Besides, we also prompt GPT-4 to generate a corresponding reason for the given score, which describes the strengths and weaknesses of the evaluated rationale from multiple perspectives.
    \item \textbf{Step 2: Expert Review} We then ask human experts familiar with the relation extraction task, to review the reason and score provided by GPT-4. The human experts may adjust the scores by increasing, maintaining, or decreasing them as necessary.
    \item \textbf{Step 3: Final Scoring} Finally, we calculate the average of the adjusted scores from the three human experts as the final quality score.
\end{itemize}

\begin{table}[t]
    \centering
    \setlength{\tabcolsep}{1mm}
    {
    \begin{tabular}{lcccc}
    \toprule
        \textbf{Evaluator} & {{\textbf{SemEval}}} & {{\textbf{TACRED}}}  & {{\textbf{Re-TACRED}}} & {\textbf{Avg.}} \\
        \midrule
        \multicolumn{5}{c}{\textit{SRVF based on Llama-2-7b-chat}} \\
        \midrule
        GPT-4          & 4.120 & 3.500 & 3.795 & 3.805 \\
        ~ w/ Human & 4.153 & 3.656 & 3.884 & 3.898 \\
        
        \midrule
        \multicolumn{5}{c}{\textit{SRVF based on GPT-3.5-turbo}} \\
        \midrule
        GPT-4          & 4.320 & 3.876 & 4.190 & 4.129 \\
        ~ w/ Human & 4.309 & 3.854 & 4.279 & 4.147 \\
    \bottomrule
    \end{tabular}
    }
    \caption{Quality evaluation on unbiased rationales generated by our proposed SRVF based on Llama-2-7b-chat and GPT-3.5-turbo. The results of human expert are average scores by three human evaluators.}
    \label{tab:quality_unbiased_rationale}
\end{table}

Table~\ref{tab:quality_unbiased_rationale} shows the evaluation results.
From Table~\ref{tab:quality_unbiased_rationale}, we can observe that:
1) The quality scores on the SemEval and Re-TACRED datasets are all around 4, i.e., ``good'' quality. This indicates that the proposed SRVF can generate high-quality unbiased rationales.
2) The quality scores on TACRED are the lowest, mainly due to its numerous annotation errors~\cite{stoica-2021-Re-TACRED}.

\begin{figure}[t]
\centering
\includegraphics[width=1.0\linewidth]{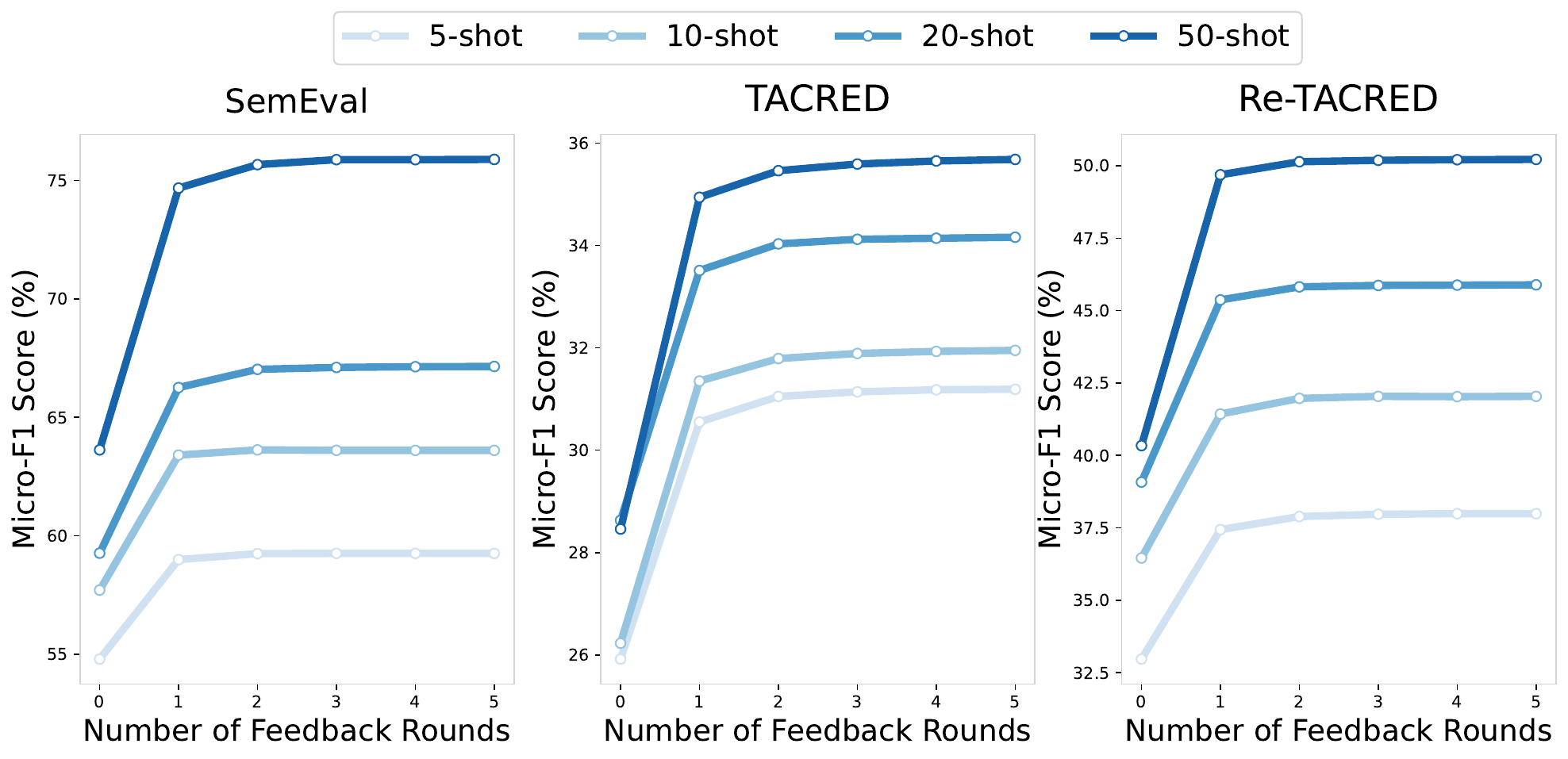}
\caption{Results (micro-F1 scores) after $k$ iterations of the verification-feedback-correction procedure. The results are averaged over the three backbones.}
\label{fig:iterative_feedback}
\end{figure}

\paragraph{Analysis of Iterative Feedback}
To explore the required iteration number in the automated feedback procedure, we visualize the performance corresponding to the number of iterations in Fig.~\ref{fig:iterative_feedback}.

From Fig.~\ref{fig:iterative_feedback}, we can find that:
1) Compared to no feedback correction, just one round of correction can bring about up to 11\% absolute F1 score improvement.
2) As the feedback rounds increase from 1 to 3, the performance continues to rise. With more than 4 feedback rounds, the performance is saturated. This indicates that our method can quickly converge to the desired state without too many iterative corrections.

\begin{figure}[t]
\centering
\includegraphics[width=0.48\textwidth]{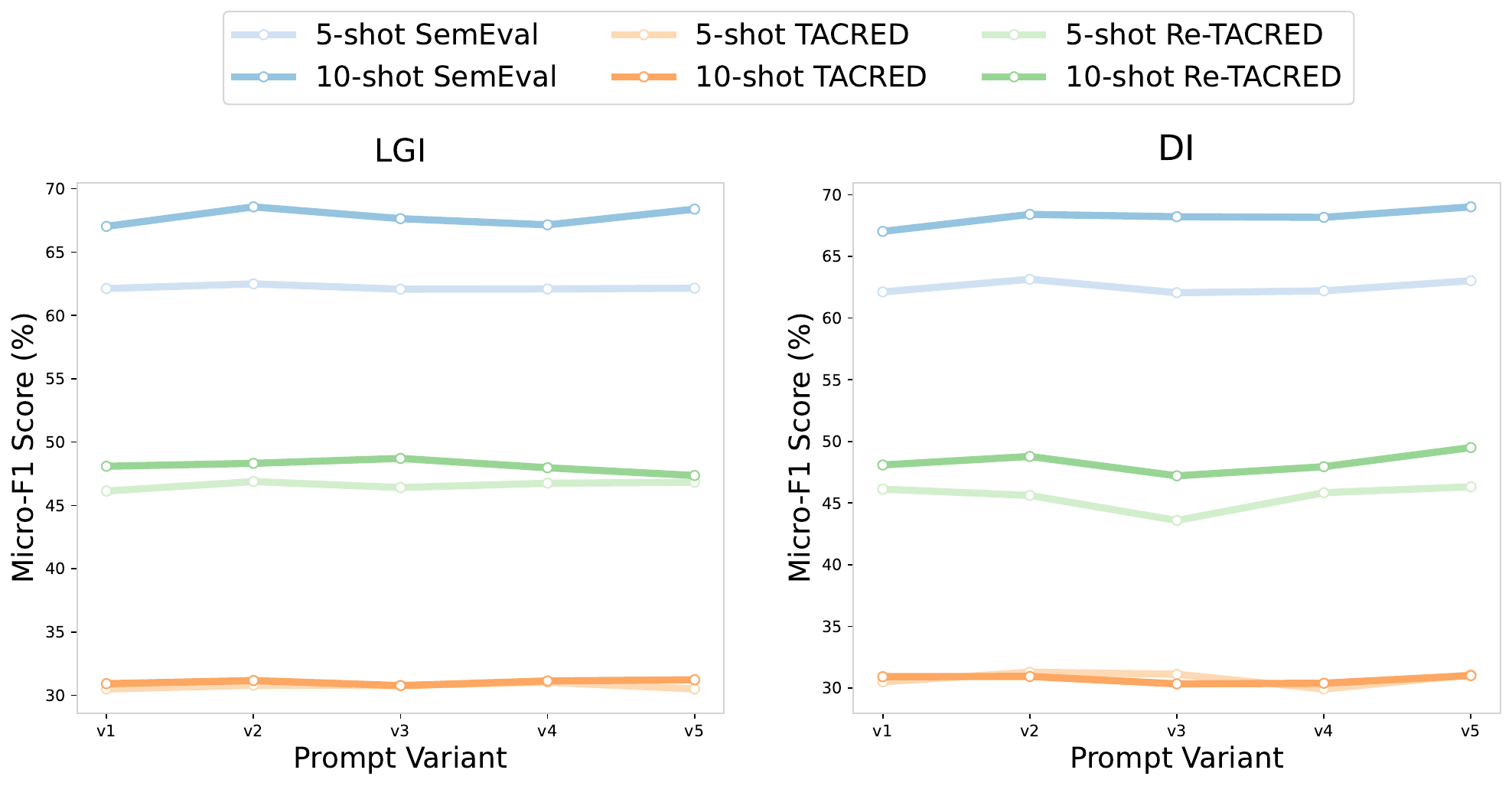}
\caption{Analysis of prompt sensitivity. Results (micro-F1 scores) are reported using Llama-2-7b-chat with the task-specific retriever.}
\label{fig:analysis_of_prompt_sensitivity}
\end{figure}

\paragraph{Analysis of Prompt Sensitivity}\label{sec:app:analysis_of_prompt_sensitivity}
To explore whether the proposed SRVF framework is sensitive to the designed prompts, we conduct an analysis of the prompt sensitivity.
Specifically, we focus on the prompts used in the proposed causal intervention and observation method (Section~\ref{sec:causal_intervention_observation}), which collects unbiased and biased rationales that are then used for training the rationale supervisor.
There are two main prompts in this method, i.e., label-guided intervention (LGI) prompt and diversified intervention (DI) prompt.
We present the designed variants of the LGI prompt and DI prompt in Table~\ref{tab:variants_for_analysis_of_prompt_sensitivity}.
The experimental results using different prompt variants are shown in Fig.~\ref{fig:analysis_of_prompt_sensitivity}.

As shown in Fig.~\ref{fig:analysis_of_prompt_sensitivity}, we can observe that there is only a small fluctuation in performance when using different prompt variants for collecting unbiased and biased rationales. 
This indicates that our proposed causal intervention and observation method is not greatly affected by prompt words or certain sentence variations, providing a guarantee for practicality and reproducibility.

\paragraph{Impact of the Hyper-parameter $\tau$}\label{sec:app:impact_of_tau_of_cl}

For the rationale contrastive training (Section~\ref{sec:rationale_supervisor_training}), we add a hyper-parameter $\tau$ to encourage the training to focus more on the hard samples, i.e., negative pairs with high similarity.
To explore whether the method is sensitive to $\tau$, we conduct experiments with different $\tau$ and show the results in Table~\ref{tab:impact_of_tau_of_cl}.

As shown in Table~\ref{tab:impact_of_tau_of_cl}, we can see that focusing too much on hard samples, i.e., $\tau \in \{0.01, 0.05\}$, leads to a significant drop in performance. 
Besides, the performance is best when $\tau = 0.50$, suggesting that a moderate focus on hard negative pairs can improve performance.
Moreover, when $\tau$ is between 0.2 and 1, the performance is relatively stable, reflecting the robustness of our method to $\tau$.

\paragraph{Impact of the Number of Feedback Demonstrations}\label{sec:app:impact_feedback_demo_num}

\begin{figure}[t!]
\centering
\includegraphics[width=0.48\textwidth]{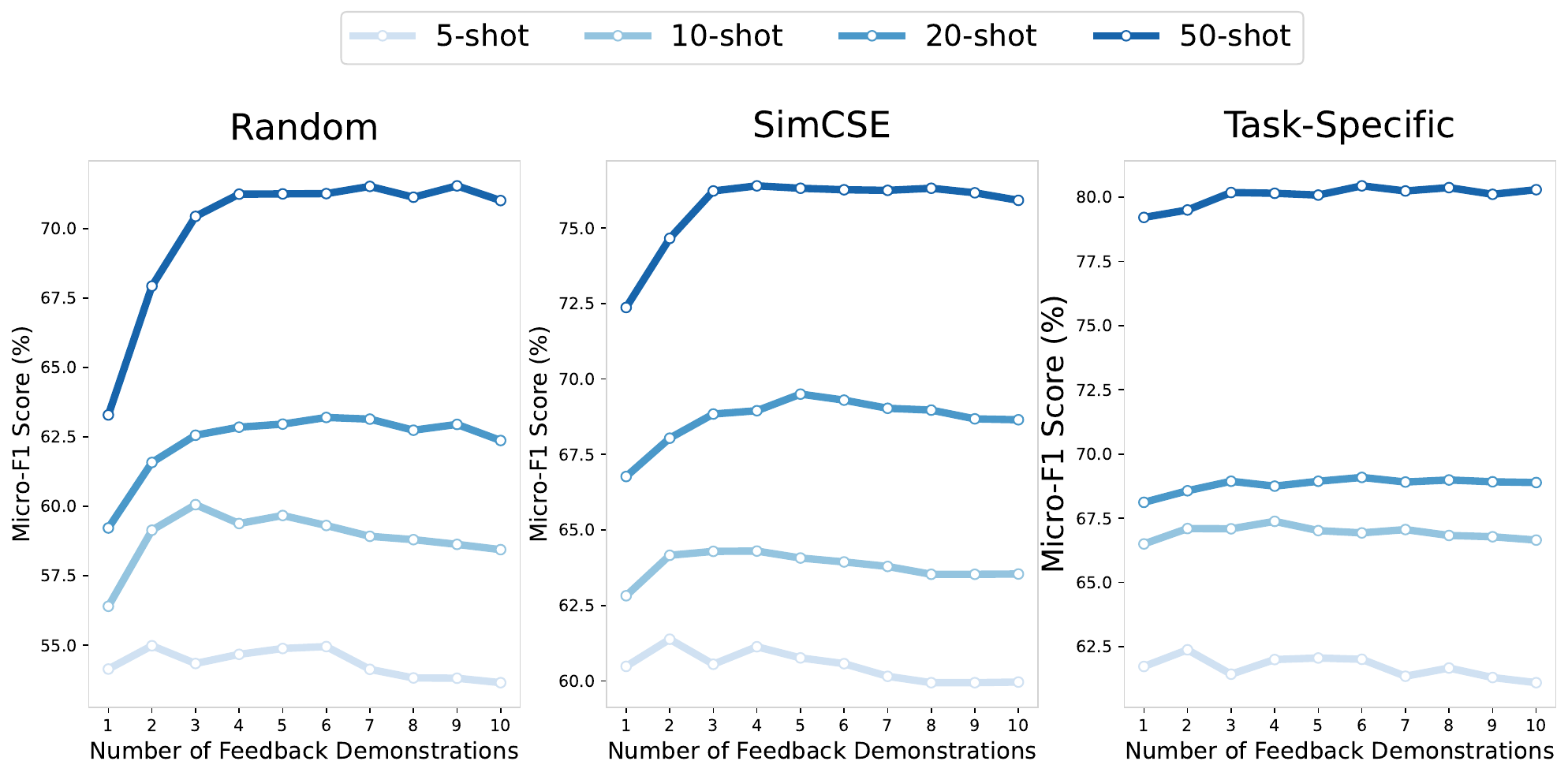}
\caption{Impact of the number of feedback demonstrations on the SemEval dataset. Results (micro-F1 scores) are reported using Llama-2-7b-chat with the three backbones.}
\label{fig:impact_of_number_of_feedback_demo}
\end{figure}

At the feedback-correction stage, too few feedback demonstrations may result in ignoring some potentially useful feedback information, while too many feedback demonstrations may introduce some noisy feedback information. 
Given this, we explore the impact of the number of feedback demonstrations on the SemEval dataset and present the results in Fig.~\ref{fig:impact_of_number_of_feedback_demo}.

\begin{table}[t]
    \centering
    \setlength{\tabcolsep}{1.5mm}
    \resizebox{\linewidth}{!}{
    \begin{tabular}{lccccccc}
    \toprule
        \multirow{2}{*}{\textbf{$\tau$}} & \multicolumn{2}{c}{\textbf{SemEval}} & \multicolumn{2}{c}{\textbf{TACRED}} & \multicolumn{2}{c}{\textbf{Re-TACRED}} & \multirow{2}{*}{\textbf{Avg.}}\\
        \cmidrule(lr){2-3}\cmidrule(lr){4-5}\cmidrule(lr){6-7}
        & {{5-shot}} & {{10-shot}}  & {{5-shot}} & {{10-shot}} & {{5-shot}} & {{10-shot}}\\
        \midrule
        0.01 & 53.28& 64.93& 26.43& 26.43& 40.09& 40.06& 35.89\\
        0.05 & 53.28& 59.21& 25.57& 26.43& 24.52& 22.07& 30.15\\
        0.10 & 60.95& 66.82& 30.55& 26.43& 46.16& 49.70& 40.09\\
        0.20 & 62.12& 67.03& 30.50& 30.92& 46.13& 48.09& 40.68\\
        0.50 & \textbf{62.22}& \textbf{69.07}& 30.22& 31.29& 47.11& \textbf{50.49}& \textbf{41.49}\\
        0.80 & 61.76& 68.74& \textbf{30.75}& 31.71& 47.24& 50.05& 41.47\\
        1.00 & 61.84& 68.50& 30.74& \textbf{32.11}& \textbf{47.25}& 49.91& 41.48\\
    \bottomrule
    \end{tabular}
    }
    \caption{Impact of the hyper-parameter $\tau$. The reported results (micro-F1 scores) are reported using Llama-2-7b-chat with the task-specific retriever. The best results are in \textbf{bold}.}
    \label{tab:impact_of_tau_of_cl}
\end{table}

As shown in Fig.~\ref{fig:impact_of_number_of_feedback_demo}, we can observe that:
1) In most cases, there is a consistent increase in performance when the number of feedback demonstrations is increased from 1 to 3, suggesting that more demonstrations provide the LLM with richer feedback information to correct the initial response.
2) When the number of feedback demonstrations is larger than 7, the performance shows a decrease, which indicates that the provided feedback demonstrations do not always play a positive role and may contain noisy ones.
This inspires future work to further explore the elimination of noise in feedback.

\begin{table*}[htp]
\centering
\setlength{\tabcolsep}{4mm}
\resizebox{\linewidth}{!}{
	\begin{tabular}{p{18cm}}
	\toprule
        \textit{Prompt}\\
        \midrule
        \textbf{Given Sentence}:  Given Sentence:  The series comprises some re-issues of the previous books , as well as new titles.\\
        \textbf{Head Entity}: titles\\
        \textbf{Tail Entity}: series\\
        \textbf{Relation}: Component-Whole\\
        \textbf{Evaluated Rationale}: In the given sentence, the key phrase "series comprises some re-issues of the previous books, as well as new titles" implies that the titles are part of the series. Therefore, the head entity "titles" serves as the "Component" while the tail entity "series" serves as the "Whole". \\

        \textbf{Task Instruction}: Please rate the quality of the explanation given for the judgments about the relationship between the head and tail entities.\\
        \textbf{Evaluation Criteria}: \\
        *Score 1 indicates very poor quality: a completely flawed explanation of the reasons why the relationships between entities constitute a specific relationship;\\
        *Score 2 indicates poor quality: some explanation of the relationship between the entities is given, but there are obvious flawed errors (e.g., the contextual keywords are not captured);\\
        *Score 3 indicates moderate quality: a partially correct explanation of the relationship between the entities, but with flawed errors;\\
        *Score 4 indicates good quality: a mostly correct explanation of the inter-entity relationship, but with very weak flaws (e.g., irrelevant grammatical issues);\\
        *Score 5 indicates very good quality: a good explanation of what constitutes a specific relationship between entities.\\
        \textbf{Score}:\\
        \textbf{Reason}: \\
        \midrule        
        \textit{Response}\\
        \midrule  
        \textbf{Score}: 4\\
        \textbf{Reason}: The explanation provided is mostly correct and clearly identifies the relationship between the head entity "titles" and the tail entity "series" as a Component-Whole relationship. The key phrase "series comprises some re-issues of the previous books, as well as new titles" is accurately interpreted to mean that the titles are part of the series. The explanation effectively captures the contextual keywords and the nature of the relationship. However, the explanation could be slightly more detailed in explaining why "titles" specifically are considered components of the "series" to achieve a perfect score.\\
	\bottomrule
	\end{tabular}
 }
 \caption{An example of the used prompt for the initial scoring step in the unbiased rationale quality evaluation.}
 \label{tab:quality_analysis_prompt_for_gpt4_evaluation}
\end{table*}

\begin{table*}[htp]
\centering
\setlength{\tabcolsep}{3mm}
\resizebox{\linewidth}{!}{
	\begin{tabular}{p{23.2cm}}
	\toprule
        \textit{Orginal (variant 1) prompt for label-guided intervention (LGI) }\\
        \midrule
        \textbf{Instruction}: Given a sentence, explain why there is certain relation between the head and tail entities in the sentence.\\ 
        \textbf{Demonstrations}: \{demonstrations\}\\ 
         Please learn the demonstration and follow the instruction, complete the "Reasoning Explanations" and "Prediction" parts of the new given instance. Please end with (End of Instance) when complete the text. \\ 
        \textbf{Inference}: \{inference\_part\}\\ 
        \midrule
        \textit{(variant 2)}\\
        \midrule
        \textbf{Instruction}: Provided a sentence, unfold the reason behind the particular relationship between the head and tail entities in the sentence.\\ 
        \textbf{Demonstrations}: \{demonstrations\}\\ 
         Please study the demonstration and adhere to the instruction, fulfill the "Reasoning Explanations" and "Prediction" areas of the new presented instance. Please finalize with (End of Instance) when you've completed the text. \\ 
        \textbf{Inference}: \{inference\_part\}\\ 
        \midrule
        \textit{(variant 3)}\\
        \midrule
        \textbf{Instruction}: Given a line of text, expound on why there exists a specific association between the head and tail entities within the sentence. \\ 
        \textbf{Demonstrations}: \{demonstrations\}\\ 
         Please understand the demonstration and go by the guideline, complete the "Reasoning Explanations" and "Prediction" sections of the fresh instance provided. Ensure to finish with (End of Instance) once the text composition is done. \\ 
        \textbf{Inference}: \{inference\_part\}\\ 
        \midrule
        \textit{(variant 4)}\\
        \midrule
        \textbf{Instruction}: Presented a sentence, illustrate why there is a defined link between the head and tail entities in the formation of the sentence. \\ 
        \textbf{Demonstrations}: \{demonstrations\}\\ 
         Please observe the demonstration and comply with the directive, finish the "Reasoning Explanations" and "Prediction" elements of the unique instance offered. Please wrap up with (End of Instance) after the text is completed. \\ 
        \textbf{Inference}: \{inference\_part\}\\ 
        \midrule
        \textit{(variant 5)}\\
        \midrule
        \textbf{Instruction}: With a referred sentence, give an explanation for the definite relation between the head and tail entities in the sentence. \\ 
        \textbf{Demonstrations}: \{demonstrations\}\\ 
         Please grasp the demonstration and follow the guideline, wrap up the "Reasoning Explanations" and "Prediction" components of the newly given instance. Please terminate with (End of Instance) upon the completion of the text. \\ 
        \textbf{Inference}: \{inference\_part\}\\ 

        \midrule
        \midrule
        \textit{Orginal (variant 1) prompt for diversified intervention (DI) }\\
        \midrule
        \textbf{Instruction}: Determine the relation between the given head entity and tail entity in the given sentence. The relation category is from the relation type set.\\ 
        \textbf{Demonstrations}: \{demonstrations\}\\ 
         Please learn the demonstration and follow the instruction, complete the "Reasoning Explanations" and "Prediction" parts of the new given instance. You only need to solve the only instance given. Please end with (End of Instance) when complete the text. \\ 
        \textbf{Inference}: \{inference\_part\}\\ 

        \midrule
        \textit{(variant 2)}\\
        \midrule
        \textbf{Instruction}: Identify the connection between the provided head entity and tail entity in the stipulated sentence. The connection category is from the relation type set.\\ 
        \textbf{Demonstrations}: \{demonstrations\}\\ 
         Please study the demonstration and adhere to the instruction, finish the "Reasoning Explanations" and "Prediction" areas of the new presented instance. You only need to tackle the single instance given. Please conclude with (End of Instance) after you have completed the text. \\ 
        \textbf{Inference}: \{inference\_part\}\\ 

        \midrule
        \textit{(variant 3)}\\
        \midrule
        \textbf{Instruction}: Ascertain the relationship between the designated head entity and tail entity in the quoted sentence. This relationship type is taken from the relation type set. 
        \textbf{Demonstrations}: \{demonstrations\}\\ 
         Please observe the demonstration and go by the guideline, complete the "Reasoning Explanations" and "Prediction" sections of the newly provided instance. You are only required to solve the only instance provided. Please finish with (End of Instance) once you have completed the composition. \\ 
        \textbf{Inference}: \{inference\_part\}\\ 

        \midrule
        \textit{(variant 4)}\\
        \midrule
        \textbf{Instruction}: Evaluate the link between the specified head entity and tail entity within the given sentence. The link group comes from the relation type set. \\ 
        \textbf{Demonstrations}: \{demonstrations\}\\ 
         Please grasp the demonstration and comply with the instruction, wrap up the "Reasoning Explanations" and "Prediction" elements of the new instance furnished. You only need to deal with the unique instance presented. Don't forget to end with (End of Instance) after finishing the text. \\ 
        \textbf{Inference}: \{inference\_part\}\\ 

        \midrule
        \textit{(variant 5)}\\
        \midrule
        \textbf{Instruction}: Figure out the association between the named head entity and tail entity in the sentence provided. The association class is derived from the relation type set. \\ 
        \textbf{Demonstrations}: \{demonstrations\}\\ 
         Please understand the demonstration and follow the directive, finalize the "Reasoning Explanations" and "Prediction" parts of the fresh instance offered. You are only required to address the sole instance given. Please terminate with (End of Instance) upon completing the text. \\ 
        \textbf{Inference}: \{inference\_part\}\\ 

	\bottomrule
	\end{tabular}
 }
 \caption{The designed variants of the LGI prompt and DI prompt. Since the \{Demonstrations\} and \{Inference\} parts are made up of specific samples, we do not design relevant prompt variants of them.}
 \label{tab:variants_for_analysis_of_prompt_sensitivity}
\end{table*}

\end{document}